\documentclass[10pt,journal,letterpaper,compsoc]{IEEEtran}
\usepackage{amsthm}
\usepackage{graphicx} 
\usepackage{supertabular}
\usepackage{slashbox}
\usepackage{makecell}
\usepackage{subfigure}
\usepackage{multirow}
\usepackage{rotating}
\usepackage{algorithm}
\usepackage{algorithmic} 
\usepackage{amsmath} 
\usepackage{amssymb}
\usepackage{bm}
\newtheorem{mydef}{Definition}
\theoremstyle{remark} 
 
\ifCLASSOPTIONcompsoc
\else
\fi
\ifCLASSINFOpdf
\else
\fi

\begin{document}
%
\title{A New Strategy of Cost-Free Learning in \\the Class Imbalance Problem}
%
%
%
%

\author{Xiaowan Zhang and
        Bao-Gang Hu,~\IEEEmembership{Senior~Member,~IEEE}   
\IEEEcompsocitemizethanks{\IEEEcompsocthanksitem X. Zhang is with the National Laboratory of Pattern Recognition, Institute of Automation Chinese Academy of Sciences, Beijing 100190, P.R. China. E-mail: xwzhang@nlpr.ia.ac.cn.

\IEEEcompsocthanksitem B.-G. Hu is with the National Laboratory of Pattern Recognition,
Institute of Automation Chinese Academy of Sciences, 95 ZhongGuanCun East Road, Beijing
100190, P.R. China. E-mail: hubg@nlpr.ia.ac.cn.
}
\thanks{}}

\IEEEcompsoctitleabstractindextext{%
\begin{abstract}
In this work, we define cost-free learning ({\bf CFL}) formally in comparison with cost-sensitive learning ({\bf CSL}). The main difference between them is that a CFL approach seeks optimal classification results without requiring any cost information, even in the class imbalance problem. In fact, several CFL approaches exist in the related studies, such as sampling and some criteria-based approaches. However, to our best knowledge, none of the existing CFL and CSL approaches are able to process the abstaining classifications properly when no information is given about errors and rejects. Based on information theory, we propose a novel CFL which seeks to maximize normalized mutual information of the targets and the decision outputs of classifiers. Using the strategy, we can deal with binary/multi-class classifications with/without abstaining. Significant features are observed from the new strategy. While the degree of class imbalance is changing, the proposed strategy is able to balance the errors and rejects accordingly and automatically. Another advantage of the strategy is its ability of deriving optimal rejection thresholds for abstaining classifications and the ``\textit{equivalent}'' costs in binary classifications. The connection between rejection thresholds and ROC curve is explored. Empirical investigation is made on several benchmark data sets in comparison with other existing approaches. The classification results demonstrate a promising perspective of the strategy in machine learning. 

\end{abstract}

}

\maketitle

\IEEEdisplaynotcompsoctitleabstractindextext

%
\IEEEpeerreviewmaketitle

\section{Introduction}
%
%

%
%
%
%
\IEEEPARstart{I}{mbalanced} data sets \cite{N. Japkowicz,Raeder T} arise frequently in a variety of real-world applications, such as medicine, biology, finance, and computer vision. Generally, users  focus more on the \textit{minority} class and consider the cost of misclassifying a minority class to be more expensive. Unfortunately, most \textit{conventional classification} algorithms assume  that the class distributions are balanced or the misclassification costs are equal. They seek to maximize the  overall accuracy which yet cannot distinguish the error types. Therefore, they may neglect the significance of the minority class and  tend toward the majority class. Learning in the class imbalance is thus of high importance in  data mining and machine learning.

 From the background of this problem, various methods are developed within a category called \textit{cost-sensitive learning} (CSL), such as costs to test \cite{Qiang:testcost}, to relabel training instances \cite{Domingos99:metacost}, to sample \cite{Zadrozny:sampling}, to weight instances \cite{Ting:weighting}, and to find a decision threshold \cite{Elkan:foundation,ShengL06:thresholdingcost}. These methods use unequal costs to make a bias toward the minority class. Generally, when the costs are not given, these methods can not work properly. A comprehensive review of learning in the class imbalance problem is provided by He and Garcia \cite{He}.

When there exist some uncertainties in the decision, it may be better to apply \textit{abstaining classification} \cite{Chow} to reduce the chance of a potential misclassification. Significant benefits have been obtained from abstaining classification, particularly in very critical applications \cite{Pietraszek1,Temanni}. The optimal rejection thresholds could be found through minimizing a loss function  in a cost-sensitive setting \cite{Landgrebe,Friedel:costcurves,Tortorella2004}. The possibility of designing loss functions for classifiers with a reject option is also explored \cite{loss}. In the context of abstaining classifications, the existing  CSL approaches require the cost terms associated to the rejects. However, one often fails to provide such information. Up to now, there seems no proper guideline to give the information in terms of the skew ratio. Obviously, a reject option adds another degree of complexity in classifications over the non-abstaining approaches. For advancing the technology and being compatible with human intelligence,      we consider the abstaining strategy will become a common option for most learning machines in future.

In the class imbalance problem, CSL is an important research direction. Based on the definition in \cite{costdefine}, we extend it below by including the situation of abstaining. 
\begin{mydef}
 \emph{Cost-Sensitive Learning (CSL)} is a type of learning that takes the misclassification costs and/or rejection costs into consideration. The goal of this type of learning is to minimize the total cost.
\end{mydef}
 CSL generally requires modelers or users to
specify cost terms for reaching the goal. However, this
work addresses one open issue which is mostly overlooked:

\emph{``How to conduct a learning in the class imbalance problem when costs are unknown for errors and rejects''?}

 In fact, the issue is not unusual in real-world applications.
Therefore, we propose another category of learning below
for distinguishing the differences between the present work and the
existing studies in CSL.

\begin{mydef}
 \emph{Cost-Free Learning (CFL)} is a type of learning that does not require the cost terms associated with the misclassifications and/or rejects as the inputs. The goal of this type of learning is to get optimal classification results without using any cost information.
\end{mydef}

\begin{figure}
\begin{center}
\includegraphics[scale=0.8]{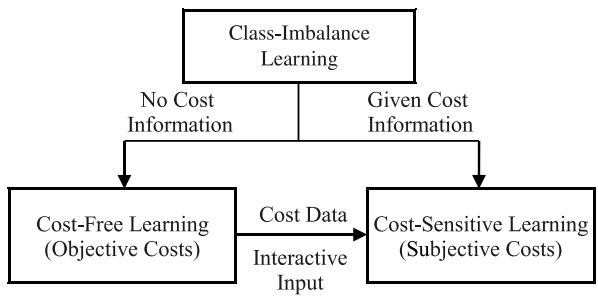}
\end{center}
\caption{Cost-Free Learning and Cost-Sensitive Learning.} \label{fig1}
\end{figure} 

It is understandable that CFL may face a bigger challenge which is shown by the fact that most existing approaches may fail to present reasonable solutions to the open issue. This work attempts to provide an applicable learning strategy in CFL.

 We extend Hu's \cite{Hu:difference} study on mutual information classifiers. While Hu presents the theoretical formulas, no learning approaches and results are shown for the real-world data sets. 
Hence, this work 
focuses on learning and presents 
main contributions as follows.
\begin{itemize}
\item We propose a CFL strategy in the class imbalance problem. Using \textit{normalized mutual information} (NI) as the learning target, we conduct the learning from cost-insensitive classifiers. Therefore, we are able to adopt conventional classifiers for simple and direct implementations. The most advantage of this strategy is its unique feature in classification scenarios
where one has no knowledge of costs.

\item We study the relations between the strategy and some existing approaches. First, we derive the ``\textit{equivalent}'' costs and the rejection thresholds for binary classifications by using the strategy. The costs, being ``\textit{objective}'' for the reason of purely determined by the distributions of the given data sets, can be a useful reference for ``\textit{subjective}'' cost specifications in CSL (Fig. \ref{fig1}). Second, we present graphical interpretations of ROC curve plots for both non-abstaining and abstaining classifiers. From the plots, the intrinsic differences between the strategy and other existing approaches are explained in the cases when one class becomes extremely rare.

\item  We conduct empirical studies on binary class and multi-class problems. Specific investigation is made on abstaining classifications, and we obtain several results from the benchmark data sets which have not been reported before in literature. The
results confirm the advantages of the strategy and show the promising perspective of CFL in imbalanced data sets.  
\end{itemize}

\subsection{Related Work}
When costs are unequal and unknown,  Maloof \cite{Maloof:unknown} uses ROC curve to  show the performance of binary classifications under different cost settings. The study can be viewed as comparing classifiers rather than finding an optimal operating point. Cost curve \cite{costcurveDrummond,Friedel:costcurves} can be used to visualize optimal expected costs over a range of cost settings, but it does not suit multi-class problem. Zadrozny and Elkan \cite{ZadroznyE01:unknown} apply least-squares multiple linear regression to estimate the costs. The method requires cost information of the training sets to be known.  Cross validation \cite{Zhou08:face} is proposed to choose from a limited set of cost values, and the final decisions are made by users.

There exists some CFL approaches in the class imbalance problem. Various sampling strategies \cite{Kubat97,SMOTE,Minlong} try to modify the imbalanced class distributions. Active learning \cite{S.Ertekin:2007} is also investigated to select desired instances and the feature selection techniques \cite{Z.Zheng, Mike10} are applied to combat the class imbalance problem for high-dimensional data sets. Besides, ensemble learning methods \cite{ensemble1,ensemble2} are used to improve the generalization of predicting the minority class. The recognition-based methods \cite{L.M,Japkowicz01} that train on a single class are proposed as alternatives to the discrimination-based methods to avoid the influence of imbalanced distributions. Sun et al. \cite{SunGmean} can get costs for multi-class data sets through maximizing the \textit{geometric mean} (G-mean)  or \textit{F-measure} which has the ability of balancing the performance of each class.  However, all CFL methods above do not take abstaining into consideration and  may fail to  process the abstaining classifications.

In regards to abstaining classification, some strategies have been proposed for defining optimal reject rules. Pietraszek \cite{Pietraszek} proposes a bounded-abstaintion model with ROC analysis,  and Fumera et al. \cite{Fumera} seek to maximize accuracy while keeping the reject rate below a given value. However, the  bound information and the targeted reject rate are required to be specified respectively. When there is no prior knowledge of these settings, it is hard to determine the values. Li and Sethi \cite{MingKunLi} restrict the maximum error rate of each class, but the rates may  conflict  when they are arbitrarily given.

\subsection{Paper Organization}
The remainder of this paper is organized as follows: In Section 2, a brief review of  NI is provided. We present our CFL strategy in Section 3.  Section 4  analyzes the relations between the optimal parameters and the cost terms, and presents the graphical interpretations of ROC curve plots. The experimental results are presented in Section 5. Finally, we conclude this work in Section 6.

\section{Review: normalized mutual information}

\begin{table}
\centering
\caption{Confusion Matrix $C$ in {\it m}-Class Abstaining Classification} \label{confusionmatrix}
\begin{tabular}{c|ccccc} \hline
  & & &Y & & \\\cline{2-6}
T &$1$ &$2$ &\ldots &$m$ &$m+1$\\
\hline
$1$ &$c_{11}$ & $c_{12}$ & \ldots & $c_{1m}$& $c_{1(m+1)}$\\
$2$ &$c_{21}$ & $c_{22}$ & \ldots & $c_{2m}$& $c_{2(m+1)}$\\
$\vdots$&$\vdots$ &$\vdots$ & $\ddots$ &$\vdots$ &$\vdots$\\
 $m$ &$c_{m1}$ & $c_{m2}$ & \ldots & $c_{mm}$& $c_{m(m+1)}$\\
\hline
\end{tabular}
\end{table}

\textit{Normalized mutual information} (NI) has been used as an evaluation criterion to measure the degree of dependence between the targets $T$ and the decision outputs $Y$, and it is denoted as 
\begin{equation}
NI(T,Y)=\frac{I(T,Y)}{H(T)},\nonumber
 \end{equation}
where $I(T,Y)$ is the mutual information of two random variables $T$ and $Y$,  $H(T)$ is the Shannon's entropy of $T$. Note that  $NI(T,Y)$ is in the range  $[0,1]$.

Suppose an $m$-class abstaining classification, with each class denoted as $1,2,\dots,m$, and the rejected class denoted as $m+1$. The value of the target variable $T$ ranges from $1$ to $m$, while the decision output variable $Y$ ranges from $1$ to $m+1$. Then we have
\begin{eqnarray}
I(T,Y)\hspace{-0.12in}&=&\hspace{-0.12in}\sum\limits_{i=1}^{m}\sum\limits_{j=1}^{m+1}
P(T=i,Y=j)\log_{2}\frac{P(T=i,Y=j)}{P(T=i)P(Y=j)},
\nonumber\\
H(T)\hspace{-0.12in}&=&\hspace{-0.12in}-\sum\limits_{i=1}^{m}P(T=i)\log_{2}P(T=i). 
\nonumber
\end{eqnarray}

 In general, as the exact probability distribution functions of $T$ and $Y$ are hard to derive, Hu et al. \cite{Hu:arxiv} apply empirical estimations to compute NI based on the confusion matrix.  Table \ref{confusionmatrix}
illustrates an augmented confusion matrix $C$ in an $m$-class abstaining classification by adding the last column as a rejected class $m+1$. The rows correspond to the states of the targets $T$, and the columns correspond to  the  states of the decision outputs $Y$.
 $c_{ij}$ represents the number of the instances that belong to the $i$-th class classified as the $j$-th class, $i=1,2,\dots,m$, $j=1,2,\dots,m+1$. To avoid unchanged value of NI if rejections are made within only one class, the formula of  NI  is proposed  as \cite{Hu:arxiv}
\begin{eqnarray}\label{NI}
NI(T,Y)&\hspace{-0.13in}=\hspace{-0.13in}&\frac{\sum\limits_{i=1}^{m}\sum\limits_{j=1}^{m}
P_{e}(T=i,Y=j)\frac{P_{e}(T=i,Y=j)}{P_{e}(T=i)P_{e}(Y=j)}}{-\sum\limits_{i=1}^{m}P_{e}(T=i)\log_{2}P_{e}(T=i)}
\nonumber\\
&\hspace{-0.1in}=\hspace{-0.1in}&-\frac{\sum\limits_{i=1}^{m}\sum\limits_{j=1}^{m}c_{ij}\log_{2}
\left(
\frac{c_{ij}}{C_{i}\sum\limits_{i=1}^{m}
\left(
\frac{c_{ij}}{n}
\right)}
\right)}{\sum\limits_{i=1}^{m}C_{i}\log_{2}(\frac{C_{i}}{n})},  
\end{eqnarray}
where $Y$ is counted from $1$ to $m$ rather than to $m+1$. The subscript ``$e$'' is given for denoting empirical terms, $C_{i}=\sum_{j=1}^{m+1}c_{ij}$ is the total number of instances in the $i$-th class, $i=1,2,\dots,m$, and  $n=\sum_{i=1}^{m}\sum_{j=1}^{m+1}c_{ij}$ is the total number in the confusion matrix.  
In \textit{non-abstaining classification}, i.e. classification without rejection,  $Y$ ranges from $1$ to $m$  and (\ref{NI}) is actually the formula  of the original NI.
Then (\ref{NI}) is applicable for both non-abstaining and abstaining classifications in the present work.

Principe et al. \cite{Principe:2000} present a schematic diagram of \textit{information theory learning} (ITL) and they mention that maximizing mutual information as the target function makes the decision outputs correlate with the targets as much as possible. Recently, a  study \cite{Hu:difference} confirms  that ITL opens a new perspective for classifier design.
MacKay \cite{Mackay} recommends mutual information for its single rankable value which makes more sense than error rate. Hu et al. \cite{Hu:arxiv}  study theoretically for the first time on both error types and reject types in binary classifications. They  consider information-theoretic measures most promising in providing ``\textit{objectivity}'' to classification evaluations in class imbalance problems.
 The above viewpoints of mutual information motivate our following NI-based strategy for CFL in the class imbalance problem.

\section{NI-Based Classification}
 In this work, we distinguish two types of classificaitons,
namely, ``{\it non-abstaining classification}'' for no 
rejection and ``{\it abstaining  classification}'' for rejection. 
From the phenomenon  that different error types and reject types produce different effects on NI, one can derive a conclusion that NI considers the costs to be unequal, unlike accuracy. In fact, the cost information is hiding in NI, and we take advantage of its bias toward the minority class.
The bias can be changed through moving the decision thresholds, and the value of NI is changed accordingly. We focus our study on the probabilistic classifiers in the present work, although it can also be applied to non-probabilistic classifiers \cite{myICDMW}.

Let $\bm{x}=[\bm{x}_{1}, \bm{x}_{2},\dots,\bm{x}_{n}]^{T}$  denote a data matrix with $n$ instances to be classified, $\bm{x}_{l} \in \mathbb{R}^{d}$ is the input feature vector, $l=1,2,\dots,n$. The target vector is denoted as $\bm{t}=[t_{1},t_{2},\dots,t_{n}]^{T}$, $t_{l}\in T=\{1,2,\dots,m\}$. The decision output vector is denoted as $ \bm{y}=[y_{1},y_{2},\dots,y_{n}]^{T}$, $y_{l}\in Y=\{1,2,\dots,m\}$ for non-abstaining classification while $y_{l}\hspace{-0.02in}\in\hspace{-0.02in} Y\hspace{-0.03in}=\hspace{-0.03in}\{1,2,\dots,m+1\}$ for abstaining  classification.
Then  for both non-abstaining and abstaining classifications, we have a generalized formula  with NI being a function of the data set and the decision thresholds:
\begin{eqnarray}\label{y_all}
NI&\hspace{-0.1in}=\hspace{-0.1in}&NI\big(\bm{t},\bm{y}=f(\bm{\varphi}(\bm{x}),\bm{\tau})\big),
\nonumber\\
y_{l}&\hspace{-0.1in}=\hspace{-0.1in}&\left\{
\begin{aligned}
\arg\max_{i}\big(\frac{\varphi_{i}(\bm{x}_{l})}{\tau_{i}}\big)  \hspace{0.2in} \mbox{if}\   \max\big(\frac{\varphi_{i}(\bm{x}_{l})}{\tau_{i}}\big)\geq 1,\\
m+1  \hspace{0.75in}  \mbox{otherwise}, \hspace{0.66in} \\
\end{aligned}\right.
\\
0&\hspace{-0.1in}<\hspace{-0.1in}&\tau_{i}\leq1,\ i=1,2,\dots,m, \ l=1,2,\dots,n,\nonumber
\end{eqnarray}
where $\bm{\varphi}(\bm{x})\in\mathbb{R}^{n\times m}$ denotes the real-value output matrix  of a probabilistic classifier for $n$ instances, $\varphi_{i}(\bm{x}_{l})$ is the probabilistic output of class $i$ for $\bm{x}_{l}$,   $\sum_{i=1}^{m}\varphi_{i}(\bm{x}_{l})\hspace{-0.04in}=\hspace{-0.04in}1$ and $0 \leq \varphi_{i}(\bm{x}_{l})\hspace{-0.04in} \leq\hspace{-0.04in} 1$. $\bm{\tau}\hspace{-0.03in}=\hspace{-0.03in}[\tau_{1},\tau_{2},\dots,\tau_{m}]^{T}\hspace{-0.03in}\in\hspace{-0.03in}\mathbb{R}^{m}$ is the vector parameter of the decision thresholds. 
The decision rule of $y_{l}$  is proposed in this form to avoid classifying an instance $\bm{x}_{l}$ into more than one class.

\subsection{Non-Abstaining Classification}
In non-abstaining classification, the first condition for deriving $y_{l}$  in (\ref{y_all}) should only be satisfied, i.e.
\begin{eqnarray}
y_{l}&\hspace{-0.1in}=\hspace{-0.1in}&\arg\max_{i}\big(\frac{\varphi_{i}(\bm{x}_{l})}{\tau_{i}}\big), 0<\tau_{i}\leq1,
\nonumber\\
i&\hspace{-0.1in}=\hspace{-0.1in}&1,2,\dots,m, l=1,2,\dots,n. \nonumber
\end{eqnarray}
Let $\phi_{i}(\bm{x}_{l})=\alpha_{i}\varphi_{i}(\bm{x}_{l})$, $\alpha_{i}$ is denoted as the  weight parameter for $\varphi_{i}(\bm{x}_{l})$, $\alpha_{i}=\frac{\tau_{m}}{\tau_{i}}$ and $\alpha_{m}=1$. Then we have the following:
\begin{eqnarray}
\phi_{i}(\bm{x}_{l})&\hspace{-0.1in}=\hspace{-0.1in}&\alpha_{i}\varphi_{i}(\bm{x}_{l})
\nonumber\\
&\hspace{-0.1in}=\hspace{-0.1in}&\tau_{m}\frac{\varphi_{i}(\bm{x}_{l})}{\tau_{i}}.\nonumber
\end{eqnarray}
It is obvious that  $\arg\max_{i}\phi_{i}(\bm{x}_{l})\hspace{-0.05in}=\hspace{-0.05in}\arg\max_{i}\big(\frac{\varphi_{i}(\bm{x}_{l})}{\tau_{i}}\big)$, and the optimal decision for $y_{l}$ remains the same.
The effect of assigning weights to the probabilistic outputs  is the same as setting decision thresholds. Therefore, we denote $\bm{\alpha}\hspace{-0.05in}=\hspace{-0.05in}[\alpha_{1},\alpha_{2},\dots,1]^{T}\hspace{-0.05in}\in\hspace{-0.05in}\mathbb{R}^{m}$ as the weight parameter vector, and the class assignment rule for $y_{l}\hspace{-0.04in}=\hspace{-0.04in}f(\bm{\varphi}(\bm{x}_{l}),\bm{\alpha})$ is based on the highest weighted probabilistic outputs.
For non-abstaining classification, we propose
\begin{eqnarray}\label{y_con}
&&\hspace{-0.15in}\mbox{maximize}\  NI\big(\bm{t},\bm{y}=f(\bm{\varphi}(\bm{x}),\bm{\alpha})\big),
\nonumber\\
&&\hspace{-0.15in}\mbox{subject to}
\nonumber\\
&&y_{l}=\arg\max_{i}\alpha_{i}\varphi_{i}(\bm{x}_{l}), \nonumber\\
&&\alpha_{i}>0, \ i=1,2,\dots,m, \ l=1,2,\dots,n.
\end{eqnarray}

In order to maximize NI, the optimal  weight parameter $\bm{\alpha^{\ast}}$ should be
\begin{equation}\label{eq:alph}
 \bm{\alpha}^{\ast}=\arg\max_{\bm{\alpha}} NI\big(\bm{t},\bm{y}=f(\bm{\varphi}(\bm{x}),\bm{\alpha})\big).
\end{equation}

\subsection{Abstaining Classification}
We denote $\bm{T_{r}}\hspace{-0.06in}=\hspace{-0.06in}[T_{r1},T_{r2},\dots,T_{rm}]^{T}\hspace{-0.05in}\in\hspace{-0.05in}\mathbb{R}^{m}$ as the rejection threshold vector in dealing with abstaining classificaiton. Let $1-T_{ri}=\tau_{i}$, 
$T_{ri}$ is in the range $[0,1)$, $i=1,2, \dots, m$. The decision output for $y_{l}=f(\bm{\varphi}(\bm{x}_{l}),\bm{T_{r}})$ lies within $m+1$ classes. Then we propose 
\begin{eqnarray}\label{y_abs}
&&\hspace{-0.15in}\mbox{maximize}\  NI\big(\bm{t},\bm{y}=f(\bm{\varphi}(\bm{x}),\bm{T_{r}})\big),
\nonumber\\
&&\hspace{-0.15in}\mbox{subject to}
\nonumber\\
&&y_{l}=\left\{
\begin{aligned}
\arg\max_{i}\big(\frac{\varphi_{i}(\bm{x}_{l})}{1-T_{ri}}\big) \hspace{0.1in}  \mbox{if} \max\big(\frac{\varphi_{i}(\bm{x}_{l})}{1-T_{ri}}\big)\geq1,\\
m+1\hspace{0.68in}   \mbox{otherwise}, \hspace{0.62in}\\
\end{aligned}\right.
\\
&&0\leq T_{ri}<1,    0\leq\sum^{m}_{i=1}T_{ri}<m-1,
\nonumber\\ 
&&i=1,2,\dots,m, \ l=1,2,\dots,n.\nonumber
\end{eqnarray}
Note that $m-1$ is the loose upper bound for the summation $\sum^{m}_{i=1}T_{ri}$. 
 Assume a situation that all instances satisfy the first condition in (\ref{y_abs}), and $\frac{\varphi_{i}(\bm{x}_{l})}{1-T_{ri}}\geq1$  for all probabilistic outputs, 
i.e. 
$\forall i,l, \ \varphi_{i}(\bm{x}_{l}) \geq 1-T_{ri}$.
Then we get the following:
\begin{eqnarray} 
\sum^{m}_{i=1}\varphi_{i}(\bm{x}_{l})&\hspace{-0.1in}\geq\hspace{-0.1in}& \sum^{m}_{i=1}(1-T_{ri}),
\nonumber\\
\sum^{m}_{i=1}T_{ri}&\hspace{-0.1in}\geq\hspace{-0.1in}& m-1.\nonumber
\end{eqnarray}
If $\sum^{m}_{i=1}T_{ri}$ falls in this interval, the condition of rejection would never be satisfied and the proposal of abstaining classification is ineffective. Reversely, this  extreme situation would not happen if $\sum^{m}_{i=1}T_{ri}< m-1$.

In order to maximize NI, the optimal rejection threshold vector $\bm{T_{r}^{\ast}}$ should be
\begin{equation}\label{eq:Tr}
 \bm{T_{r}}^{\ast}=\arg\max_{\bm{T_{r}}} NI\big(\bm{t},\bm{y}=f(\bm{\varphi}(\bm{x}),\bm{T_{r}})\big).
\end{equation}

\subsection{Optimization Algorithm}
 The present framework is proposed based on the confusion matrix from which we compute  NI, but it is not differentiable. We apply a general optimization algorithm called  ``\textit{Powell Algorithm}''  which is a direct method for nonlinear optimization without calculating the derivatives \cite{Powell:Powell}. It is also widely used in image registration to find optimal registration parameters.
\begin{algorithm}[htb]\footnotesize
\caption{Learning algorithm} 
\label{alg:alg1}  
\begin{algorithmic}[1]
\REQUIRE  
Probabilistic  outputs $\bm{\varphi}(\bm{x})$, target labels $\bm{t}$, $\mathcal{D}$ as the degree of freedom in $\bm{\tau}$.
\ENSURE
$\bm{\tau^{\ast}}$\\
\STATE Initialize $\bm{\tau_{1}}$ as a random vector in the range of $\bm{\tau}$, $\bm{d_{1}},\bm{d_{2}},\dots, \bm{d_{\mathcal{D}}}$ as linear independent vectors, 
 number of iterations $W=0$, $\varepsilon \geq 0$.
\STATE \textbf{Iterative Search Phase:}
\STATE \textbf{repeat}
\STATE \hspace{0.1in}$W=W+1$. Let $\bm{\tau^{(1)}_{W}}=\bm{\tau_{W}}$.\label{codestart}
\STATE \hspace{0.1in}\textbf{for}  {each direction $\bm{d_{i}}$,  $i = 1$  \textbf{to}  $\mathcal{D}$} \textbf{do}\label{for}
  \STATE  \hspace{0.2in} $\bar{\eta}^{(i)}=\arg\max_{\eta \in \mathbb{R}}NI\big(\bm{t}, \bm{y}=f(\bm{\varphi}(\bm{x}), \bm{\tau^{(i)}_{W}}+\eta\bm{ d_{i}})\big)$;\label{min1} \\
  \STATE \hspace{0.2in} Update $\bm{\tau_{W}}$ in the current direction: $\bm{\tau^{(i+1)}_{W}} =  \bm{\tau^{(i)}_{W}}+ \bar{\eta}^{(i)}\bm{d_{i}}$; \\ \label{forend}
\STATE \hspace{0.06in} \textbf{end for}
  \STATE \hspace{0.1in}Update the directions: $\bm{d_{i}}=\bm{ d_{i+1}}, i= 1,2,\dots,\mathcal{D}-1$;\\ $\hspace{1.24in}\bm{d_{\mathcal{D}}}=\bm{\tau^{(\mathcal{D}+1)}_{W}}-\bm{\tau_{W}}$;\\
 \STATE \hspace{0.1in}$\eta^{\ast}_{W}=\arg\max_{\eta \in \mathbb{R}}NI\big(\bm{t}, \bm{y}=f(\bm{\varphi}(\bm{x}),\bm{\tau_{W}}+\eta\bm{ d_{\mathcal{D}}})\big)$;\label{min2}\\
 \STATE \hspace{0.1in}Update $\bm{\tau}$ after the current iteration:  $\bm{\tau_{W+1}} = \bm{ \tau_{W}}+\eta^{\ast}_{W}\bm{d_{\mathcal{D}}}$;
\STATE \textbf{until} $ || \bm{\tau_{W+1}}-\bm{\tau_{W}} ||_{2} \leq \varepsilon$
\STATE Return $\bm{\tau^{\ast}} = \bm{\tau_{W+1}} $.
\end{algorithmic}
\end{algorithm}

The algorithm is given in Algorithm \ref{alg:alg1}, which we apply to find $\bm{\tau^{\ast}}$ for demonstration.  We can also apply it to both $\bm{\alpha}$ and $\bm{T_{r}}$. For Step \ref{min1} and Step \ref{min2}, we use {\it bracketing method} to find three starting points and use {\it Brent's Method} to realize one-dimensional optimization.
$W$ iterations of the basic procedure lead to $W(\mathcal{D}+1)$ one-dimensional optimizations.
 One disadvantage of this algorithm is that it may find a local extrema. Hence, we randomly choose the starting points several times and then pick the best one.
In non-abstaining binary  classification,  $\mathcal{D}=1$,  so we just work from Step \ref{codestart} to Step \ref{forend} once and assign the value of $\bm{\tau^{(2)}_{W}}$ to $\bm{\tau^{\ast}}$.

\section{Relations in Binary Classification }
The previous section completes the essence of the present framework.
It can be regarded as a generic   way to make the conventional learning algorithms information-theoretic based. 

The optimal parameters reflect the degree of bias implied by NI, and may reveal the cost information to some extent.
In this section, we focus on binary classification and analyze the relations between the optimal parameters and the cost terms. 
Moreover, we discover some graphical interpretations of performance measures on ROC curve, which allows the users to adjust the parameters more conveniently using ROC curve.

\subsection{Normalized Cost Matrix}
Friedel et al. \cite{Friedel:costcurves} derive normalized cost matrix based on the \textit{overall  risk} which is written  as
\begin{eqnarray}
Risk=\sum_{i,j}\lambda_{ij}p(j|i)p(i),
\end{eqnarray}
where $\lambda_{ij}$ is the original cost in the common cost matrix that assigns an instance of  class $i$ to class $j$, $p(j|i)$ is the true probability in such situation, and $p(i)$ is the true prior probability of class $i$.
The \textit{conditional risk} of assigning an instance $\bm{x}_{l}$ to class $j$  is 
\begin{eqnarray}
Risk(j|\bm{x}_{l})=\sum^{m}_{i=1}\lambda_{ij}p(i|\bm{x}_{l}),
\end{eqnarray}
where $p(i|\bm{x}_{l})$ is the true posterior probability of  class $i$ given $\bm{x}_{l}$. By applying the way of transforming costs \cite{Friedel:costcurves}, we find that the normalization way for the overall risk is also applicable for the conditional risk.

In binary classification, we refer to class $1$ and class $2$ as \textit{negative class} ($N$) and \textit{positive class} ($P$), respectively.
We denote $\lambda_{FN}$, $\lambda_{FP}$, $\lambda_{TN}$,  $\lambda_{TP}$, $\lambda_{RN}$ and $\lambda_{RP}$
 to be the costs of \textit{false negative}, \textit{false positive}, \textit{true negative}, \textit{true positive}, \textit{reject negative}, and \textit{reject positive}, respectively.  
  Therefore, the normalized cost matrix for non-abstaining binary classification  can be denoted as
\begin{eqnarray}\label{eq:conven_cost}
\bar{\lambda}_{no\_rej} =
\left[ \begin{array}{cc}
\bar{\lambda}_{TN} & \bar{\lambda}_{FP} \\
\bar{\lambda}_{FN} & \bar{\lambda}_{TP} \\
\end{array} \right]=
\left[ \begin{array}{cc}
0 & \bar{\lambda}_{FP} \\
1 &0 \\ 
\end{array}\right]
\end{eqnarray}
with $\beta=\lambda_{FN}-\lambda_{TP}$, then $\bar{\lambda}_{TN}=\frac{\lambda_{TN}-\lambda_{TN}}{\beta}=0$, $\bar{\lambda}_{FP}=\frac{\lambda_{FP}-\lambda_{TN}}{\beta}$, $\bar{\lambda}_{FN}=\frac{\lambda_{FN}-\lambda_{TP}}{\beta}=1$, $\bar{\lambda}_{TP}=\frac{\lambda_{TP}-\lambda_{TP}}{\beta}=0$.

Similarly, the normalized cost matrix for abstaining binary classification  can be denoted as
\begin{eqnarray}\label{eq:abs_cost}
\bar{\lambda}_{rej} =
\left[ \begin{array}{ccc}
\bar{\lambda}_{TN} & \bar{\lambda}_{FP} &\bar{\lambda}_{RN} \\
\bar{\lambda}_{FN} & \bar{\lambda}_{TP} & \bar{\lambda}_{RP} \\
\end{array} \right]=
\left[ \begin{array}{ccc}
0 & \bar{\lambda}_{FP} &\bar{\lambda}_{RN} \\
1 & 0& \bar{\lambda}_{RP} \\
\end{array} \right]
\end{eqnarray}
with $\bar{\lambda}_{TN}=0, \bar{\lambda}_{FP}=\frac{\lambda_{FP}-\lambda_{TN}}{\beta}, \bar{\lambda}_{RN}=\frac{\lambda_{RN}-\lambda_{TN}}{\beta}, \bar{\lambda}_{FN}=1, \bar{\lambda}_{TP}=0, \bar{\lambda}_{RP}=\frac{\lambda_{RP}-\lambda_{TP}}{\beta}, \beta=\lambda_{FN}-\lambda_{TP}$.
 The first two columns contain the misclassification costs, while the last column indicates the rejection costs.

It is reasonable to assume that the values of the original 
 correct classification costs and misclassification costs in
 the common cost matrix are not affected by introducing
 a reject option. Therefore, what is noteworthy is that $\bar{\lambda}_{FP}$ in (\ref{eq:conven_cost}) is consistent with that in  (\ref{eq:abs_cost}).

\subsection{Optimal Weight  and Misclassification Cost}
In  non-abstaining binary classification, it is feasible to set the decision thresholds as $\bm{\tau}=[1-\tau_{P}, \tau_{P}]^{T}$, which has one degree of freedom.

 The relation between the decision thresholds and the costs has been  derived by Elkan \cite{Elkan:foundation} through  minimizing the conditional risk. Considering the normalized cost matrix in (\ref{eq:conven_cost}), the decision threshold $\tau^{\ast}_{P}$ of the positive class for making optimal decision can be represented as
\begin{eqnarray}\label{Elkan_t}
\tau^{\ast}_{P}=\frac{\bar{\lambda}_{FP}}{1+\bar{\lambda}_{FP}},
\end{eqnarray}
with $\bar{\lambda}_{FP}$  be the variable. It is required that the value of $\bar{\lambda}_{FP}$  be given and be reasonable. Otherwise,  $\tau^{\ast}_{P}$ can not be derived or not be proper.

In our present work, the optimal weight vector is $\bm{\alpha}^{\ast}=[\alpha^{\ast}_{N},1]^{T}$. We apply it in the decision rule of maximum weighted posterior probability. Then the optimal prediction is the positive class if and only if $\alpha^{\ast}_{N}p(N|\bm{x}_{l})\leq p(P|\bm{x}_{l})$.
Hence, the decision threshold $\tau^{\ast\ast}_{P}$ of the positive class for making optimal decision is 
\begin{eqnarray}\label{my_t}
\tau^{\ast\ast}_{P}=\frac{\alpha^{\ast}_{N}}{1+\alpha^{\ast}_{N}}.
\end{eqnarray}

Suppose that the minimum conditional risk rule shares the same decision thresholds with the maximum weighted posterior probability rule, then  (\ref{Elkan_t}) and (\ref{my_t}) should be equal.  And  we give the following definition:
\begin{mydef}
Given the optimal weight $\alpha^{\ast}_{N}$,  the \emph{``equivalent'' misclassification cost} is defined as
\begin{eqnarray}\label{lambda}
\bar{\lambda}_{FP}&\hspace{-0.1in}=\hspace{-0.1in}&\alpha^{\ast}_{N}.
\end{eqnarray}
\end{mydef}

 In general, it is assumed that $\bar{\lambda}_{FP}<\bar{\lambda}_{FN}$, i.e. $\bar{\lambda}_{FP}<1$. In this case, it is required that  $\alpha^{\ast}_{N}<1$.

\begin{figure*}[ht]
\centering
\mbox{
\subfigure[For non-abstaining classification]{\includegraphics[scale=0.9]{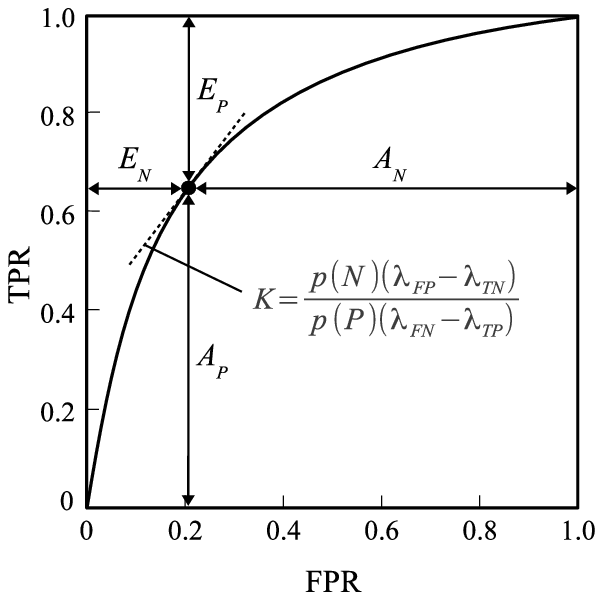}} \hspace{0.25in}
\subfigure[For abstaining classification]{\includegraphics[scale=0.9]{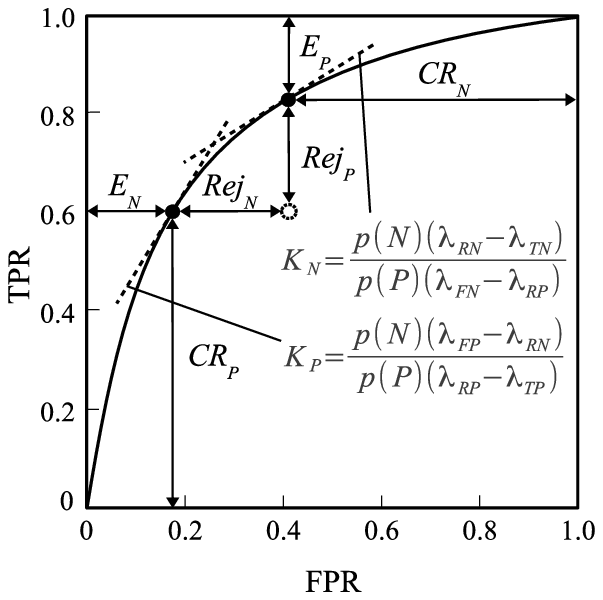}} 
}
\caption{Graphical interpretations  of ROC curves. (a) For non-abstaining classification. (b) For abstaining classification.} \label{fig2}
\end{figure*}

\subsection{Optimal Rejection Thresholds and Costs}
In abstaining  binary classification,  the  relations between the rejection thresholds and the costs can be presented in a form of explicit formulae \cite{Hu:difference}. With the optimal rejection threshold vector $\mathbf{T^{\ast}_{r}}=[T^{\ast}_{rN}, T^{\ast}_{rP}]^{T}$ and the normalized cost matrix in (\ref{eq:abs_cost}), these relations are 
\begin{eqnarray}\label{eq:Tcost}
&&T^{\ast}_{rN} = \frac{\bar{\lambda}_{RN}}{1+\bar{\lambda}_{RN}-\bar{\lambda}_{RP}},
\nonumber\\
&&T^{\ast}_{rP} = \frac{\bar{\lambda}_{RP}}{\bar{\lambda}_{FP}-\bar{\lambda}_{RN}+\bar{\lambda}_{RP}}, \nonumber
\end{eqnarray}
which imply a parameter redundancy.
In addition, the value of $\bar{\lambda}_{FP}$ derived from (\ref{lambda}) can be utilized as a prior knowledge under the assumption of cost consistency. 
\begin{mydef}
Given the ``\textit{equivalent}'' misclassification cost  $\bar{\lambda}_{FP}=\alpha^{\ast}_{N}$,  the \emph{``equivalent'' rejection costs} are defined as 
\begin{eqnarray}\label{eq:rejcost}
\bar{\lambda}_{RN}&\hspace{-0.1in}=\hspace{-0.1in}& \frac{T^{\ast}_{rN}(1-T^{\ast}_{rP})-T^{\ast}_{rN}T^{\ast}_{rP}\bar{\lambda}_{FP}}{1-T^{\ast}_{rN}-T^{\ast}_{rP}},
\nonumber\\
\bar{\lambda}_{RP}&\hspace{-0.1in}=\hspace{-0.1in}& \frac{-T^{\ast}_{rN}T^{\ast}_{rP}+(1-T^{\ast}_{rN})T^{\ast}_{rP}\bar{\lambda}_{FP}}{1-T^{\ast}_{rN}-T^{\ast}_{rP}}.
\end{eqnarray}
\end{mydef}
Based on \cite{Hu:difference}, one can have the relations $\bar{\lambda}_{TN} < \bar{\lambda}_{RN}<\bar{\lambda}_{FP}$ and  $\bar{\lambda}_{TP}< \bar{\lambda}_{RP}<\bar{\lambda}_{FN}$. Then we can  obtain the following properties from (\ref{eq:rejcost}):
\begin{enumerate}
\renewcommand{\labelenumi}{P\theenumi.}
\item If $0< \bar{\lambda}_{RN}<\bar{\lambda}_{FP}$, we have $0<T^{\ast}_{rN}<\frac{\alpha^{\ast}_{N}}{1+\alpha^{\ast}_{N}}$ and $T^{\ast}_{rP}<\frac{1}{1+\alpha^{\ast}_{N}}$;
\item If $0< \bar{\lambda}_{RP}<1$, we have $T^{\ast}_{rN}<\frac{\alpha^{\ast}_{N}}{1+\alpha^{\ast}_{N}}$ and $0<T^{\ast}_{rP}<\frac{1}{1+\alpha^{\ast}_{N}}$;
\item If  $0<T^{\ast}_{rN}<\frac{\alpha^{\ast}_{N}}{1+\alpha^{\ast}_{N}}$ and $0<T^{\ast}_{rP}<\frac{1}{1+\alpha^{\ast}_{N}}$, then $0< \bar{\lambda}_{RN}<\bar{\lambda}_{FP}$ and $0< \bar{\lambda}_{RP}<1$.
\end{enumerate}

\subsection{Graphical Interpretations of ROC Curve Plots with/without Abstaining  }
In binary classification, an ROC curve plot presents complete information about the performance of each class \cite{Graphical2011}, so that an \textit{overall performance measure} \cite{Hu2008}, such as AUC, can be formed. This is a preferred feature in processing class imbalance problems \cite{Fawcett2006}. Furthermore, an ROC curve can also provide the graphical interpretations for non-abstaining and abstaining classifications in Fig. \ref{fig2}, where TPR and FPR are {\it true positive rate} and {\it false positive rate}. We denote $\textit{A}$, $\textit{CR}$, $\textit{E}$ and $\textit{Rej}$ to be {\it accuracy}, {\it correct recognition rate}, {\it error rate}, and {\it reject rate}, respectively. $C_{N}$ and $C_{P}$ are the total numbers of the negatives and positives, respectively.
$C_{FN},C_{FP},C_{TN},C_{TP},C_{RN}$ and $C_{RP}$ are the numbers of the \textit{false negatives}, \textit{false positives}, \textit{true negatives}, \textit{true positives}, \textit{reject negatives}, and \textit{reject positives},   respectively.  
 Their relations are shown as follows:

\begin{subequations}
\hspace{-0.14in}Non-abstaining:
\begin{eqnarray}
&&\hspace{-0.5in}A_{N}+ E_{N}=1,  \text{and} \  A_{P}+E_{P}=1,
\nonumber\\
&&\hspace{-0.5in}A_{N}=\frac{C_{TN}}{C_{N}}, E_{N}=\frac{C_{FP}}{C_{N}}, A_{P}=\frac{C_{TP}}{C_{P}}, E_{P}=\frac{C_{FN}}{C_{P}};
\end{eqnarray}
Abstaining:
\begin{eqnarray}
&&\hspace{-0.5in}CR_{N}+E_{N}+Rej_{N}=1, \text{and}\  CR_{P}+E_{P}+Rej_{P}=1, 
\nonumber\\
&&\hspace{-0.5in}CR_{N}=\frac{C_{TN}}{C_{N}}, E_{N}=\frac{C_{FP}}{C_{N}}, Rej_{N}=\frac{C_{RN}}{C_{N}},
\nonumber\\
&&\hspace{-0.5in}CR_{P}=\frac{C_{TP}}{C_{P}}, E_{P}=\frac{C_{FN}}{C_{P}}, Rej_{P}=\frac{C_{RP}}{C_{P}}.
\end{eqnarray}
\end{subequations}

Several observations are summarized below for understanding the features of ROC plots. To begin with, we discuss an ROC curve in a non-abstaining classification, as shown in Fig. \ref{fig2}a. For a theoretical ROC curve which is concave, the decision is made by $\textit{K}$, the \textit{slope of ROC curve}, in the form of \cite{Provost and Fawcett2001}:
\begin{eqnarray}\label{eq:K}
\textit{K}=\frac{p(\textit{N})}{p(\textit{P})}\frac{\lambda_{FP}-\lambda_{TN}}{\lambda_{FN}-\lambda_{TP}}=\frac{p(\textit{N})}{p(\textit{P})}\bar{\lambda}_{FP},
\end{eqnarray}
which is also equivalent to the likelihood ratio \cite{Duda}:
\begin{eqnarray}
\textit{L}=\frac{p(x|\textit{P})}{p(x|\textit{N})}=\frac{p(\textit{N})}{p(\textit{P})}\frac{\lambda_{FP}-\lambda_{TN}}{\lambda_{FN}-\lambda_{TP}}=\frac{p(\textit{N})}{p(\textit{P})}\bar{\lambda}_{FP}.
\end{eqnarray}

From (\ref{eq:K}), one can observe that:
\begin{subequations}\label{eq18}
\begin{equation}\label{eq:18a}
\hspace{-1in}if\  p(\textit{P})\to 0, then\  \textit{K}\to \infty, 
\end{equation}
\begin{equation}\label{eq:18b}
and \ E_{P}=1, A_{P}=0, E_{N}=0, A_{N}=1,
\end{equation}
\end{subequations}
for general cost terms. (\ref{eq:18a}) indicates that the tangent point on the ROC curve will be located at  the origin in Fig. \ref{fig2}a, and (\ref{eq:18b}) demonstrates a graphical interpretation why conventional classifiers fail to process minority class (herein the positive class) properly. However, the situation in (\ref{eq18}) can never appear from using the present strategy, because it will result in a zero value of mutual information \cite{Mackay,Hu2008}.

Different with the non-abstaining classification, Fig. \ref{fig2}b shows the abstaining classification graphically on an ROC curve. Two \textit{abstaining slopes}, \textit{$K_{N}$} and \textit{$K_{P}$}, are generally given in the forms of \cite{Tortorella2004}:
\begin{eqnarray}\label{eq:knkp}
K_{N}&\hspace{-0.1in}=\hspace{-0.1in}&\frac{p(N)}{p(P)}\frac{\lambda_{RN}-\lambda_{TN}}{\lambda_{FN}-\lambda_{RP}}=\frac{p(N)}{p(P)}\frac{\bar{\lambda}_{RN}}{1-\bar{\lambda}_{RP}},
\nonumber\\
K_{P}&\hspace{-0.1in}=\hspace{-0.1in}&\frac{p(N)}{p(P)}\frac{\lambda_{FP}-\lambda_{RN}}{\lambda_{RP}-\lambda_{TP}}=\frac{p(N)}{p(P)}\frac{\bar{\lambda}_{FP}-\bar{\lambda}_{RN}}{\bar{\lambda}_{RP}}.
\end{eqnarray}

Whenever \textit{$K_{N}\neq K_{P}$}, one can observe the non-zero results of rejection rates. (\ref{eq:knkp}) confirms the finding in \cite{Hu:difference} that at most two independent parameters will determine the rejection range in binary classifications. Sometimes, one can still apply a single independent parameter, such as \textit{$K_{P}=2K_{N}$}, for abstaining decisions.

 There exist relations between rejection thresholds in the \textit{posterior curve plot} \cite{Hu:difference}   and abstaining slopes in the \textit{ROC curve plot}. Their relations and the associated constraint are derived from \cite{Hu:difference}:
\begin{eqnarray}\label{eq:knkpTr}
K_{N}=\frac{p(N)}{p(P)}\frac{T_{rN}}{1-T_{rN}}&\hspace{-0.15in},\hspace{-0.1in}& K_{P}=\frac{p(N)}{p(P)}\frac{1-T_{rP}}{T_{rP}},
\nonumber\\
K_{N}&\hspace{-0.1in}<\hspace{-0.1in}& K_{P}.
\end{eqnarray}

\section{Experiments}

\begin{table}[!t]\scriptsize
\begin{center}
\caption{Description of the Data Sets} \label{dataset}
\begin{tabular}{|l|r|r|c|l|} \hline
Data Set & \#Inst &\#Attr &\#C&Class Distribution\\
 \hline
\hline
Ism&11,180&7&2&10,920/260(=42.00)\\
Nursery(very\_recom)&12,960&9&2&12,632/328(=38.51)\\
Letter(A)&20,000&17&2&19,211/789(=24.35)\\
Rooftop&17,829&10&2&17,048/781(=21.83)\\
Pendigits(5)&10,992&17&2&9,937/1,055(=9.42)\\
Optdigits(8)&5,620&65&2&5,066/554(=9.14)\\
Vehicle(opel)&846&19&2&634/212(=2.99)\\
Yeast(NUC)&1,484&10&2&1,055/429(=2.46)\\
Phoneme&5,404&6&2&3,818/1,586(=2.41)\\
German Credit&1,000&25&2&700/300(=2.33)\\
Diabetes&768&9&2&500/268(=1.87)\\
Gamma&19,020&11&2&12,332/6,688(=1.84)\\
\hline
Cardiotocography&2,126&22&3&1,655/295/176\\
Thyroid&7,200&22&3&6,666/368/166\\
Car&1,728&7&4&1,210/384/65/69\\
Pageblock&5,473&11&5&4,913/329/28/88/115\\
\hline
\end{tabular}
\end{center}
\footnotesize
(\#Inst: number of instances, \#Attr: number of attributes, \#C:  number of classes)
\end{table}

\subsection{Configuration}
Table \ref{dataset} lists twelve binary class and four multi-class data sets  with imbalanced class distributions. 
 On \textit{Pageblock}, the maximum ratio between the majority class and the minority class is $175$.
Most of the data sets are obtained from the  UCI Machine Learning Repository\footnote{http://archive.ics.uci.edu/ml/}, 
\textit{Ism} is from \cite{SMOTE}, \textit{Rooftop} is from \cite{Maloof:unknown}, and \textit{Phoneme} is from KEEL Datasets\footnote{http://sci2s.ugr.es/keel/datasets.php}.
 All of them have continuous attributes and are rescaled to be in the range $[0,1]$.
We perform 3-fold cross validation and all experiments are repeated ten times to get the average results. In addition, Table \ref{process} lists the procedure of  our NI based experiments for each run.

We call our NI based non-abstaining classification and NI based abstaining classification \textit{``NI\_no\_rej''} and \textit{``NI\_rej''} respectively. To illustrate the effectiveness of our strategy, we adopt \textit{$k$NN} and \textit{Bayes classifier} as the conventional classifiers, and we compare our methods with \textit{SMOTE}, \textit{Cost-sensitive learning}, \textit{Chow's reject}  \cite{Chow} methods and the G-mean based methods (\textit{``Gmean\_no\_rej''} and \textit{``Gmean\_rej''}) besides two conventional classifications.

In \textit{$k$NN classifier}, we apply Euclidean distance and use the confidence values \cite{Arlandis,myICDMW} as the probabilistic outputs. The class assignment is decided by the highest confidence. For brevity, 
we just list the results of $11$-NN on all data sets except $5$-NN on \textit{Pageblock}.
In \textit{Bayes classifier}, we derive the estimated class-conditional density from the Parzen-window estimation with
 Gaussian kernel \cite{Tong00restrictedbayes} and apply Bayes rule to classification. The smooth parameter is chosen as the average value of the distance from one instance to its $r$th nearest neighborhood  ($r$$=$$10$ empirically), and the empirical probability of the occurrence of class is chosen as  the prior probability.
In \textit{SMOTE}, the average results are presented with the amount from $1$ to $5$, and it performs simultaneously  on the minority classes of the multi-class data sets  with the same amount.
 In \textit{Cost-sensitive learning}, we simply assign the inverse of the class distribution ratio to the misclassification cost $\lambda_{ij}$ for  $i$$\neq$$j$, and $\lambda_{ii}$$=$$0$. 
We do not consider abstaining for it because the rejection costs would be hard to give.
In \textit{Chow's reject}, we simply assign 0.3 to the rejection thresholds for all classes.
 In G-mean based methods, we apply our way of parameter settings and optimization to maximize G-mean.

\begin{table}\scriptsize
\centering
\caption{The Procedure of Our NI Based Experiments} \label{process}
\begin{tabular}{|p{8.3cm}|} \hline
1. Apply $3$-fold cross validation on a data set. $\frac{2}{3}$ data belong to the training\\
\ \ \   set and the remainder belong to the test set.\\
\ \  a. Apply $3$-fold cross validation on the training set. $\frac{2}{3}$ data belong to the\\
\ \ \ \ estimation set and the remainder belong to the validation set.\\
\ \ \ \ \ \ i. Apply Algorithm \ref{alg:alg1} several times to get the best parameter in each \\
\ \ \ \ \ \ \ cross validation.\\
\ \  b. Apply the mean value of 3 best parameters in step a to the training set.\\
\ \  c. Predict the test set with the parameter obtained from step b.\\
2. Obtain the results of 3 test sets.\\
\hline
\end{tabular}
\end{table}

\subsection{Evaluation Criterion}
In order to show the changes of each class clearly, \textit{E$_{i}$} and \textit{Rej$_{i}$} are applied as the error rate and the reject rate within its $i$th class respectively. The total error rate (\textit{``E''})
 and the total reject rate (\textit{``Rej''})
 are also applied. \textit{``A''} is short for the total accuracy. \textit{``G''} is short for G-mean with the formula $G$$-$$mean$$=$$(\prod_{i=1}^{m}A_{i})^{\frac{1}{m}}$, where $A_{i}$ represents the accuracy within its $i$th class.
In binary class tasks, we  also evaluate F-measure (\textit{``F''} for short).

\begin{table*}[htbp]\scriptsize
\begin{center}
\caption{ Evaluation Results on Binary Class Data Sets, ``--'': Not Available, The Best Performance in Each Cell is Bolded} \label{result1}

\begin{tabular}{|l|c|l|c|c|c|c|c|c|c|c|c|c|}
 \hline
Data set & \multicolumn{2}{c|}{Method} &$E_{N}$(\%) & $E_{P}$(\%)& $E$(\%) & $A$(\%)  &  $Rej_{N}$(\%)  & $Rej_{P}$(\%) & $Rej$(\%) & $G$(\%)&$F$(\%)  & $NI$\\
\hline 
\multirow{16}{*}{Ism} &\multirow{8}{0.1in}{$k$\\N\\N}& $k$NN classifier&$0.27$ &$49.03$ &$1.40$ &$98.60$ &--- &---  &--- &$71.22$ &$62.71$ &$0.3763$ \\
{}&{} &SMOTE&$1.13$ &$33.17$ &$1.87$ &$98.13$  &--- &--- &--- &$81.13$ &$62.66$ &$0.4361$ \\
{}&{}& Cost-sensitive&$4.66$ &$17.33$ &$4.96$ &$95.04$ &--- &--- &---&$88.74$ &$43.78$ &$0.4101$ \\
{}&{}& Gmean\_no\_rej&$4.11$  &$17.61$ & $4.42$ &$95.58$ &--- &--- &---&$88.87$ &$46.57$ &$0.4251$ \\
{}&{} &NI\_no\_rej &$1.13$ &$29.94$ &$1.80$ &$98.20$ &--- &--- &---&$83.17$ &$64.69$ &$0.4612$ \\
{}&{} &Chow's reject&$\bm{0.11}$ &$37.84$ &$\bm{0.99}$ &$\bm{99.00}$ &$0.45$ &$23.93$ &$0.99$ &$70.86$ &$64.31$ &$0.3260$\\
{}&{} &Gmean\_rej&$2.72$ &$\bm{17.54}$ &$3.06$ &$96.88$ &$1.91$ &$4.62$ &$1.97$ &$\bm{89.04}$ &$54.50$ &$0.4553$\\
{}&{} &NI\_rej&$0.82$ &$18.85$ &$1.24$ &$98.71$ &$3.10$ &$15.88$ &$3.40$ &$87.67$ &$\bm{71.04}$ &$\bm{0.4815}$\\\cline{2-13}
{}&\multirow{8}{0.1in}{B\\a\\y\\e\\s}& Bayes classifier&$0.23$ &$50.28$ &$1.40$ &$98.60$ &--- &---  &--- &$70.34$&$62.20$ &$0.3707$ \\
{}&{} &SMOTE&$0.77$ &$36.37$ &$1.60$ &$98.40$ &--- &---  &--- &$79.27$ &$64.79$ &$0.4353$ \\
{}&{} & Cost-sensitive&$12.38$ &$12.77$ &$12.39$ &$87.61$ &--- &---  &--- &$87.40$ &$24.74$ &$0.2960$ \\
{}&{}& Gmean\_no\_rej&$7.50$ &$16.68$ &$7.71$ &$92.29$ &--- &---  &--- &$87.73$&$35.99$ &$0.3567$ \\
{}&{} &NI\_no\_rej&$1.27$ &$28.15$ &$1.90$ &$98.10$ &--- &---  &--- &$84.18$&$63.82$ &$0.4664$ \\
{}&{} &Chow's reject&$\bm{0.08}$  &$39.92$ &$\bm{1.01}$  &$\bm{98.98}$ &$0.44$ &$19.09$ &$0.87$ &$71.05$ &$65.22$ &$0.3489$ \\
{}&{} &Gmean\_rej&$1.60$  &$\bm{9.37}$ &$1.78$  &$96.63$ &$41.53$ &$18.42$ &$40.99$&$\bm{92.61}$ &$68.39$ &$0.4761$ \\
{}&{} &NI\_rej&$1.10$ &$16.08$ &$1.44$ &$98.45$ &$6.82$ &$12.96$ &$6.96$ &$89.74$ &$\bm{69.71}$ &$\bm{0.5011}$\\\cline{2-13}
\hline 
\multirow{16}{*}{Nursery} &\multirow{8}{0.1in}{$k$\\N\\N} & $k$NN classifier&$\bm{0.02}$ &$38.75$ &$1.00$ &$99.00$ &--- &---  &--- &$78.23$&$75.56$ &$0.5270$ \\
{}&{} &SMOTE&$1.45$ &$12.09$ &$1.72$ &$98.28$  &--- &--- &--- &$92.83$ &$72.86$ &$0.6415$ \\
{}&{}& Cost-sensitive&$5.79$ &$\bm{0.61}$ &$5.66$ &$94.34$ &--- &--- &---&$96.76$ &$47.10$ &$0.5614$ \\
{}&{}& Gmean\_no\_rej& $2.83$ &$1.89$ &$2.81$  &$97.19$ &--- &--- &---&$97.63$ &$64.03$ &$0.6608$ \\
{}&{} &NI\_no\_rej &$0.79$ &$9.39$ &$1.00$ &$99.00$ &--- &--- &---&$94.80$ &$82.11$ &$0.7154$ \\
{}&{} &Chow's reject&$\bm{0.02}$ &$11.71$ &$0.31$ &$99.68$ &$0.36$ &$56.82$ &$1.79$ &$85.31$ &$83.53$ &$0.3753$\\
{}&{} &Gmean\_rej&$0.16$ &$0.73$ &$\bm{0.17}$ &$\bm{99.82}$ &$4.86$ &$36.90$ &$5.67$ &$\bm{99.38}$ &$\bm{96.15}$ &$0.7450$\\
{}&{} &NI\_rej&$0.57$ &$2.10$ &$0.61$ &$99.37$ &$2.53$ &$9.72$ &$2.71$ &$98.53$ &$88.23$ &$\bm{0.7686}$\\\cline{2-13}
{}&\multirow{8}{0.1in}{B\\a\\y\\e\\s}& Bayes classifier&$\bm{0.00}$ &$100.00$ &$2.52$ &$97.48$ &--- &---  &--- &$0.00$&$0.00$ &$0.0000$ \\
{}&{} &SMOTE&$0.04$ &$85.42$ &$2.20$ &$97.80$ &--- &---  &--- &$28.03$ &$21.64$ &$0.1117$ \\
{}&{} & Cost-sensitive&$29.95$ &$\bm{0.00}$ &$29.19$ &$70.81$ &--- &---  &--- &$83.69$ &$14.78$ &$0.2523$ \\
{}&{}& Gmean\_no\_rej&$7.14$ &$2.56$ &$7.03$ &$92.97$ &--- &---  &--- &$95.10$&$41.72$ &$0.5002$ \\
{}&{} &NI\_no\_rej&$3.24$ &$12.29$ &$3.47$ &$96.53$ &--- &---  &--- &$92.05$&$56.90$ &$0.5219$ \\
{}&{} &Chow's reject&$\bm{0.00}$  &$98.84$ &$2.50$  &$97.50$ &$0.01$ &$1.16$ &$0.04$&$0.00$ &$0.00$ &$0.0000$ \\
{}&{} &Gmean\_rej& $0.04$ &$0.12$ &$\bm{0.04}$  &$\bm{99.95}$ &$16.28$ &$79.58$ &$17.88$ &$\bm{99.67}$ &$\bm{95.98}$ &$0.4817$ \\
{}&{} &NI\_rej&$1.52$ &$0.88$ &$1.50$ &$98.35$ &$8.69$ &$23.32$ &$9.06$ &$98.61$ &$72.71$ &$\bm{0.5890}$\\\cline{2-12}
\hline 
\multirow{16}{*}{Letter} &\multirow{8}{0.1in}{$k$\\N\\N} & $k$NN classifier&$0.02$ &$4.21$ &$0.19$ &$99.81$ &--- &---  &--- &$97.86$&$97.56$ &$0.9192$ \\
{}&{} &SMOTE&$0.17$ &$1.03$ &$0.20$&$99.80$  &--- &--- &--- &$99.40$ &$97.47$ &$0.9418$ \\
{}&{}& Cost-sensitive&$1.18$ &$\bm{0.00}$ &$1.13$ &$98.87$ &--- &--- &---&$99.41$ &$87.44$ &$0.8379$ \\
{}&{}& Gmean\_no\_rej& $0.52$ &$0.18$ &$0.51$  &$99.49$ &--- &--- &---&$99.65$ &$93.99$ &$0.9029$ \\
{}&{} &NI\_no\_rej &$0.13$ &$1.14$ &$0.17$ &$99.83$ &--- &--- &---&$99.36$ &$97.91$ &$0.9452$ \\
{}&{} &Chow's reject&$\bm{0.01}$ &$1.39$ &$0.06$ &$99.94$ &$0.09$ &$6.62$ &$0.35$ &$99.25$ &$99.15$ &$0.9189$\\
{}&{} &Gmean\_rej&$\bm{0.01}$ &$0.05$ &$\bm{0.02}$ &$\bm{99.98}$ &$0.83$ &$8.14$ &$1.12$ &$\bm{99.97}$ &$\bm{99.79}$ &$0.9561$\\
{}&{}&NI\_rej&$0.06$ &$0.22$ &$0.07$ &$99.93$ &$0.42$ &$2.45$ &$0.50$ &$99.86$ &$99.16$ &$\bm{0.9585}$\\\cline{2-13}
{}&\multirow{8}{0.1in}{B\\a\\y\\e\\s}& Bayes classifier&$\bm{0.00}$ &$100.00$ &$3.95$ &$96.05$ &--- &---  &--- &$0.00$&$0.00$ &$0.0000$ \\
{}&{} &SMOTE&$0.02$ &$84.76$ &$3.36$ &$96.64$ &--- &---  &--- &$24.65$ &$20.72$ &$0.1229$ \\
{}&{} & Cost-sensitive&$14.56$ &$\bm{0.00}$ &$13.99$ &$86.01$ &--- &---  &--- &$92.43$ &$36.12$ &$0.4318$ \\
{}&{}& Gmean\_no\_rej&$2.14$ &$12.34$ &$2.54$ &$97.46$ &--- &---  &--- &$92.61$&$73.31$ &$0.5996$ \\
{}&{} &NI\_no\_rej&$0.78$ &$14.87$ &$1.34$ &$98.66$ &--- &---  &--- &$91.90$&$83.41$ &$0.6687$ \\
{}&{} &Chow's reject&$\bm{0.00}$  &$14.04$ &$0.55$  &$99.44$ &$0.11$ &$43.65$ &$1.83$&$86.65$ &$85.76$ &$0.4598$ \\
{}&{} &Gmean\_rej& $0.01$ &$0.03$ &$\bm{0.01}$  &$\bm{99.96}$ &$59.64$ &$90.98$ &$60.88$&$\bm{99.76}$ &$\bm{98.80}$ &$0.1784$ \\
{}&{} &NI\_rej&$0.49$ &$11.18$ &$0.91$ &$99.05$ &$5.12$ &$7.92$ &$5.23$ &$93.52$ &$87.66$ &$\bm{0.6780}$\\\cline{2-12}
\hline 
\multirow{16}{*}{Rooftop} &\multirow{8}{0.1in}{$k$\\N\\N} & $k$NN classifier&$0.59$ &$76.99.$ &$3.94$ &$96.06$ &--- &---  &--- &$47.78$&$33.82$ &$0.1264$ \\
{}&{} &SMOTE&$4.20$ &$54.28$ &$6.39$ &$93.61$  &--- &--- &--- &$65.36$ &$38.40$ &$0.1779$ \\
{}&{}& Cost-sensitive&$19.12$ &$21.03$ &$19.20$ &$80.80$ &--- &--- &---&$79.90$ &$26.48$ &$0.1926$ \\
{}&{}& Gmean\_no\_rej&$18.53$  &$21.25$ &$18.65$  &$81.35$ &--- &--- &---&$80.07$ &$27.07$ &$0.1964$ \\
{}&{} &NI\_no\_rej &$8.46$ &$37.99$ &$9.75$ &$90.25$ &--- &--- &---&$75.28$ &$35.97$ &$0.2027$ \\
{}&{} &Chow's reject&$\bm{0.18}$ &$63.71$ &$\bm{2.97}$ &$\bm{96.96}$ &$1.57$ &$23.56$ &$2.54$ &$40.66$ &$27.23$ &$0.0917$\\
{}&{} &Gmean\_rej&$8.30$ &$\bm{20.92}$ &$8.85$ &$90.08$ &$10.63$ &$17.62$ &$10.94$ &$\bm{82.19}$ &$38.09$ &$0.2345$\\
{}&{} &NI\_rej&$3.36$ &$24.67$ &$4.29$ &$94.97$ &$13.69$ &$28.84$ &$14.35$ &$79.45$ &$\bm{48.85}$ &$\bm{0.2388}$\\\cline{2-13}
{}&\multirow{8}{0.1in}{B\\a\\y\\e\\s}& Bayes classifier&$0.35$ &$81.68$ &$3.92$ &$96.08$ &--- &---  &--- &$42.64$&$29.01$ &$0.1063$ \\
{}&{} &SMOTE&$1.89$ &$63.31$ &$4.58$ &$95.42$ &--- &---  &--- &$59.19$ &$40.37$ &$0.1744$ \\
{}&{} & Cost-sensitive&$17.33$ &$19.74$ &$17.43$ &$82.57$ &--- &---  &--- &$81.45$ &$28.77$ &$0.2170$ \\
{}&{}& Gmean\_no\_rej&$17.53$ &$19.00$ &$17.59$ &$82.41$ &--- &---  &--- &$81.71$&$28.82$ &$0.2009$ \\
{}&{} &NI\_no\_rej&$10.27$ &$33.14$ &$11.27$ &$88.73$ &--- &---  &--- &$76.96$&$35.64$ &$0.2176$ \\
{}&{} &Chow's reject& $\bm{0.06}$ &$67.37$ &$3.01$  &$\bm{96.92}$ &$1.10$ &$25.56$ &$2.17$&$30.67$ &$17.01$ &$0.0587$ \\
{}&{} &Gmean\_rej&$1.30$  &$\bm{2.25}$ &$\bm{1.34}$  &$95.15$ &$71.17$ &$66.75$ &$70.98$&$\bm{94.25}$ &$\bm{69.59}$ &$0.1895$ \\
{}&{} &NI\_rej&$3.53$ &$12.02$ &$3.90$ &$94.70$ &$25.92$ &$38.73$ &$26.48$ &$87.57$ &$53.07$ &$\bm{0.2679}$\\\cline{2-12}
\hline 
\multirow{16}{*}{Pendigits} &\multirow{8}{0.1in}{$k$\\N\\N} & $k$NN classifier&$0.16$ &$1.01$ &$0.24$&$99.76$ &--- &---  &--- &$99.41$&$98.73$ &$0.9525$ \\
{}&{} &SMOTE&$0.28$ &$0.50$ &$0.30$ &$99.70$  &--- &--- &--- &$99.61$ &$98.45$ &$0.9509$ \\
{}&{}& Cost-sensitive&$0.55$ &$0.34$ &$0.53$ &$99.47$ &--- &--- &---&$99.55$ &$97.29$ &$0.9284$ \\
{}&{}& Gmean\_no\_rej& $0.31$ &$0.46$ &$0.33$  &$99.67$ &--- &--- &---&$99.62$ &$98.32$ &$0.9482$ \\
{}&{} &NI\_no\_rej &$0.24$ &$0.59$ &$0.27$ &$99.73$ &--- &--- &---&$99.59$ &$98.60$ &$0.9536$ \\
{}&{} &Chow's reject&$\bm{0.08}$ &$0.63$ &$0.14$ &$99.86$ &$0.18$ &$1.10$ &$0.27$ &$99.64$ &$99.28$ &$0.9625$\\
{}&{} &Gmean\_rej&$\bm{0.08}$ &$\bm{0.27}$ &$\bm{0.09}$ &$\bm{99.90}$ &$0.69$ &$3.47$ &$0.96$ &$\bm{99.82}$ &$\bm{99.50}$ &$0.9635$\\
{}&{} &NI\_rej&$0.10$ &$0.40$ &$0.13$ &$99.87$ &$0.32$ &$1.08$ &$0.40$ &$99.75$ &$99.31$ &$\bm{0.9655}$\\\cline{2-13}
{}&\multirow{8}{0.1in}{B\\a\\y\\e\\s}& Bayes classifier&$0.14$ &$9.01$ &$0.99$ &$99.01$ &--- &---  &--- &$95.32$&$94.62$ &$0.8245$ \\
{}&{} &SMOTE&$1.09$ &$2.64$ &$1.24$ &$98.76$ &--- &---  &--- &$98.11$ &$93.86$ &$0.8536$ \\
{}&{} & Cost-sensitive&$4.89$ &$0.15$ &$4.43$ &$95.57$ &--- &---  &--- &$97.45$ &$81.22$ &$0.7200$ \\
{}&{}& Gmean\_no\_rej&$1.11$ &$0.89$ &$1.09$ &$98.91$ &--- &---  &--- &$99.00$&$94.60$ &$0.8746$ \\
{}&{} &NI\_no\_rej&$0.62$ &$2.27$ &$0.78$ &$99.22$ &--- &---  &--- &$98.55$&$96.01$ &$0.8843$ \\
{}&{} &Chow's reject& $\bm{0.02}$ &$2.07$ &$0.22$  &$99.78$ &$0.55$ &$22.75$ &$2.68$&$98.64$ &$98.52$ &$0.7887$ \\
{}&{} &Gmean\_rej&$0.03$  &$\bm{0.06}$ &$\bm{0.03}$  &$\bm{99.96}$ &$5.80$ &$27.52$ &$7.89$ &$\bm{99.94}$ &$\bm{99.78}$ &$0.8414$ \\
{}&{} &NI\_rej&$0.36$ &$0.52$ &$0.38$ &$99.61$ &$2.58$ &$4.10$ &$2.72$ &$99.54$ &$97.98$ &$\bm{0.9061}$\\
\hline

\hline 
\multicolumn{13}{r}{\small\sl (Continued on next page)}\\
\end{tabular}
\end{center}

\end{table*}

\begin{table*}[htbp]\scriptsize
\begin{center}
\begin{tabular}{|l|c|l|c|c|c|c|c|c|c|c|c|c|}
\multicolumn{13}{l}{\small\sl (Continued from previous page)}\\
 \hline
Data set & \multicolumn{2}{c|}{Method} &$E_{N}$(\%) & $E_{P}$(\%)& $E$(\%) & $A$(\%)  &  $Rej_{N}$(\%)  & $Rej_{P}$(\%) & $Rej$(\%) & $G$(\%)&$F$(\%)  & $NI$\\
\hline 
\multirow{16}{*}{Optdigits}&\multirow{8}{0.1in}{$k$\\N\\N} & $k$NN classifier&$0.10$ &$4.80$ &$0.56$ &$99.44$ &--- &---  &--- &$97.52$&$97.08$ &$0.8938$ \\
{}&{} &SMOTE&$0.59$ &$1.65$ &$0.69$ &$99.31$  &--- &--- &--- &$98.88$ &$96.57$ &$0.9019$ \\
{}&{}& Cost-sensitive&$1.09$ &$1.19$ &$1.10$ &$98.90$ &--- &--- &---&$98.86$ &$94.65$ &$0.8709$ \\
{}&{}& Gmean\_no\_rej&$0.84$  &$1.34$ &$0.89$  &$99.11$ &--- &--- &---&$98.91$ &$95.64$ &$0.8868$ \\
{}&{} &NI\_no\_rej &$0.28$ &$2.26$ &$0.48$ &$99.52$ &--- &--- &---&$98.72$ &$97.59$ &$0.9160$ \\
{}&{} &Chow's reject&$\bm{0.06}$ &$2.35$ &$0.28$ &$99.72$ &$0.21$ &$7.29$ &$0.90$ &$98.70$ &$98.45$ &$0.8898$\\
{}&{} &Gmean\_rej&$0.07$ &$\bm{0.76}$ &$\bm{0.14}$ &$\bm{99.86}$ &$1.85$ &$10.06$ &$2.66$ &$\bm{99.54}$ &$\bm{99.25}$ &$0.9221$\\
{}&{} &NI\_rej&$0.15$ &$1.41$ &$0.27$ &$99.72$ &$0.74$ &$2.04$ &$0.87$ &$99.21$ &$98.58$ &$\bm{0.9307}$\\\cline{2-13}
{}&\multirow{8}{0.1in}{B\\a\\y\\e\\s}& Bayes classifier&$\bm{0.00}$ &$100.00$ &$9.86$ &$90.14$ &--- &---  &--- &$0.00$&$0.00$ &$0.0000$ \\
{}&{} &SMOTE&$1.29$ &$60.59$ &$7.14$ &$92.86$ &--- &---  &--- &$46.68$ &$40.41$ &$0.2741$ \\
{}&{} & Cost-sensitive&$52.14$ &$\bm{0.00}$ &$47.00$ &$53.00$ &--- &---  &--- &$69.18$ &$29.56$ &$0.1855$ \\
{}&{}& Gmean\_no\_rej&$6.19$ &$3.90$ &$5.96$ &$94.04$ &--- &---  &--- &$94.94$&$76.13$ &$0.6109$ \\
{}&{} &NI\_no\_rej&$4.39$ &$8.26$ &$4.77$ &$95.23$ &--- &---  &--- &$93.62$&$79.18$ &$0.6134$ \\
{}&{} &Chow's reject&$\bm{0.00}$  &$100.00$ &$9.86$  &$90.14$ &$0.00$ &$0.00$ &$0.00$&$0.00$ &$0.00$ &$0.0000$ \\
{}&{} &Gmean\_rej&$0.08$  &$0.18$ &$\bm{0.09}$  &$\bm{99.86}$&$39.84$ &$59.48$ &$41.78$&$\bm{99.70}$&$\bm{99.02}$&$0.4653$ \\
{}&{} &NI\_rej&$2.50$ &$1.53$ &$2.40$ &$97.31$ &$10.44$ &$15.21$ &$10.91$ &$97.69$ &$87.48$ &$\bm{0.6626}$\\\cline{2-12}
\hline 
\multirow{16}{*}{Vehicle}&\multirow{8}{0.1in}{$k$\\N\\N} & $k$NN classifier&$5.81$ &$68.30$ &$21.47$ &$78.53$ &--- &---  &--- &$54.46$&$42.35$ &$0.0915$ \\
{}&{} &SMOTE&$30.69$ &$24.93$ &$29.25$ &$70.75$  &--- &--- &--- &$70.47$ &$55.69$ &$0.1542$ \\
{}&{}& Cost-sensitive&$21.33$ &$34.90$ &$24.73$ &$75.27$ &--- &--- &---&$71.45$ &$56.95$ &$0.1442$ \\
{}&{}& Gmean\_no\_rej&$29.47$  &$24.50$ &$28.23$  &$71.77$ &--- &--- &---&$72.79$ &$57.28$ &$0.1523$ \\
{}&{} &NI\_no\_rej &$39.57$ &$14.94$ &$33.40$ &$66.60$ &--- &--- &---&$71.14$ &$56.21$ &$0.1579$ \\
{}&{} &Chow's reject&$\bm{0.79}$ &$36.43$ &$9.72$ &$86.61$ &$19.09$ &$52.44$ &$27.44$ &$47.72$ &$36.05$ &$0.1039$\\
{}&{} &Gmean\_rej&$3.97$ &$\bm{3.49}$ &$\bm{3.85}$ &$\bm{90.71}$ &$55.62$ &$74.26$ &$60.29$ &$\bm{88.49}$ &$\bm{74.85}$ &$0.1764$\\
{}&{} &NI\_rej&$14.67$ &$6.86$ &$12.71$ &$80.69$ &$34.63$ &$42.54$ &$36.61$ &$82.46$ &$68.11$ &$\bm{0.1982}$\\\cline{2-13}
{}&\multirow{8}{0.1in}{B\\a\\y\\e\\s}& Bayes classifier&$0.62$ &$93.13$ &$23.80$ &$76.20$ &--- &---  &--- &$23.22$&$11.94$ &$0.0288$ \\
{}&{} &SMOTE&$30.04$ &$41.96$ &$33.03$ &$66.97$ &--- &---  &--- &$57.28$ &$42.86$ &$0.0755$ \\
{}&{} & Cost-sensitive&$26.97$ &$40.42$ &$30.33$ &$69.67$ &--- &---  &--- &$65.86$ &$49.54$ &$0.0790$ \\
{}&{}& Gmean\_no\_rej &$36.91$ &$25.71$ &$34.09$ &$65.91$ &--- &---  &--- &$68.24$&$52.08$ &$0.1034$ \\
{}&{} &NI\_no\_rej&$45.48$ &$17.48$ &$38.46$ &$61.54$ &--- &---  &--- &$66.52$&$51.85$ &$0.1072$ \\
{}&{} &Chow's reject& $\bm{0.00}$ &$48.42$ &$12.13$  &$82.85$ &$21.86$ &$51.58$ &$29.31$&$0.00$ &$0.00$ &$0.0236$ \\
{}&{} &Gmean\_rej& $2.40$ &$\bm{1.60}$ &$\bm{2.20}$  &$\bm{91.40}$ &$73.51$ &$82.10$ &$75.66$&$\bm{84.71}$ &$\bm{73.28}$ &$0.1223$ \\
{}&{} &NI\_rej&$9.64$ &$8.30$ &$9.30$ &$81.23$ &$50.04$ &$58.16$ &$52.07$ &$79.80$ &$65.67$ &$\bm{0.1322}$\\\cline{2-12}
\hline 
\multirow{16}{*}{Yeast}&\multirow{8}{0.1in}{$k$\\N\\N} & $k$NN classifier&$9.44$ &$58.16$ &$23.52$ &$76.48$ &--- &---  &--- &$61.50$&$50.64$ &$0.1086$ \\
{}&{} &SMOTE&$36.11$ &$25.29$ &$32.98$ &$67.02$  &--- &--- &--- &$67.48$ &$56.39$ &$0.1180$ \\
{}&{}& Cost-sensitive&$25.08$ &$32.82$ &$27.32$ &$72.68$ &--- &--- &---&$70.92$ &$58.70$ &$0.1289$ \\
{}&{}& Gmean\_no\_rej&$30.99$  &$26.62$ &$29.73$  &$70.27$ &--- &--- &---&$71.02$ &$58.84$ &$0.1289$ \\
{}&{} &NI\_no\_rej &$31.52$ &$26.25$ &$30.00$ &$70.00$ &--- &--- &---&$70.35$ &$58.72$ &$0.1340$ \\
{}&{} &Chow's reject&$\bm{3.07}$ &$34.17$ &$12.06$ &$83.31$ &$20.68$ &$45.41$ &$27.83$ &$59.93$ &$49.45$ &$0.1108$\\
{}&{} &Gmean\_rej&$3.45$ &$\bm{3.45}$ &$\bm{3.45}$ &$\bm{89.39}$ &$65.17$ &$74.78$ &$67.95$ &$\bm{87.59}$ &$\bm{77.92}$ &$0.1421$\\
{}&{}&NI\_rej&$13.20$ &$9.86$ &$12.23$ &$79.35$ &$41.55$ &$41.96$ &$41.67$ &$80.30$ &$69.91$ &$\bm{0.1607}$\\\cline{2-13}
{}&\multirow{8}{0.1in}{B\\a\\y\\e\\s}& Bayes classifier&$1.60$ &$89.32$ &$26.96$ &$73.04$ &--- &---  &--- &$32.23$&$18.55$ &$0.0313$ \\
{}&{} &SMOTE&$50.67$ &$26.09$ &$43.57$ &$56.43$ &--- &---  &--- &$50.16$ &$45.92$ &$0.0719$ \\
{}&{} & Cost-sensitive&$51.30$ &$10.35$ &$39.46$ &$60.54$ &--- &---  &--- &$65.99$ &$56.81$ &$0.1234$ \\
{}&{}& Gmean\_no\_rej&$30.48$ &$30.12$ &$30.38$ &$69.62$ &--- &---  &--- &$69.38$&$57.19$ &$0.1138$ \\
{}&{} &NI\_no\_rej&$35.56$ &$26.50$ &$32.93$ &$67.07$ &--- &---  &--- &$66.49$&$55.84$ &$0.1202$ \\
{}&{} &Chow's reject&$\bm{0.19}$  &$20.33$ &$6.01$  &$\bm{88.02}$ &$38.44$ &$78.60$ &$50.05$&$19.78$ &$8.77$ &$0.0742$ \\
{}&{} &Gmean\_rej&$2.67$  &$\bm{3.12}$ &$\bm{2.80}$  &$86.14$ &$77.69$ &$81.72$ &$78.86$&$\bm{83.67}$ &$\bm{73.48}$ &$0.0822$ \\
{}&{} &NI\_rej&$12.15$ &$9.46$ &$11.37$ &$80.98$ &$40.91$ &$41.45$ &$41.06$ &$81.53$ &$71.72$ &$\bm{0.1786}$\\\cline{2-12}
\hline 
\multirow{16}{*}{Phoneme}&\multirow{8}{0.1in}{$k$\\N\\N} & $k$NN classifier&$6.44$ &$23.03$ &$11.31$ &$88.69$ &--- &---  &--- &$84.86$&$79.98$ &$0.4261$ \\
{}&{} &SMOTE&$17.40$ &$10.46$ &$15.36$ &$84.64$  &--- &--- &--- &$85.81$ &$77.51$ &$0.4106$ \\
{}&{}& Cost-sensitive&$13.69$ &$11.55$ &$13.06$ &$86.94$ &--- &--- &---&$87.37$ &$79.90$ &$0.4372$ \\
{}&{}& Gmean\_no\_rej&$14.09$  &$11.00$ &$13.18$  &$86.82$ &--- &--- &---&$87.43$ &$79.86$ &$0.4388$ \\
{}&{} &NI\_no\_rej &$12.53$ &$12.77$ &$12.60$ &$87.40$ &--- &--- &---&$87.32$ &$80.28$ &$0.4406$ \\
{}&{} &Chow's reject&$2.27$ &$11.80$ &$5.07$ &$93.94$ &$11.08$ &$29.17$ &$16.39$ &$90.10$ &$87.21$ &$0.4683$\\
{}&{} &Gmean\_rej&$\bm{1.20}$ &$\bm{1.94}$ &$\bm{1.42}$ &$\bm{97.62}$ &$36.01$ &$52.78$ &$40.93$ &$\bm{96.93}$ &$\bm{94.98}$ &$0.4509$\\
{}&{} &NI\_rej&$5.35$ &$4.62$ &$5.14$ &$93.54$ &$20.22$ &$21.79$ &$20.68$ &$93.69$ &$89.41$ &$\bm{0.5086}$\\\cline{2-13}
{}&\multirow{8}{0.1in}{B\\a\\y\\e\\s}& Bayes classifier&$10.44$ &$31.25$ &$16.55$ &$83.45$ &--- &---  &--- &$78.46$&$70.92$ &$0.2816$ \\
{}&{} &SMOTE&$24.61$ &$12.53$ &$21.06$ &$78.94$ &--- &---  &--- &$80.70$ &$71.02$ &$0.3122$ \\
{}&{} & Cost-sensitive&$21.89$ &$12.41$ &$19.10$ &$80.90$ &--- &---  &--- &$82.71$ &$72.91$ &$0.3248$ \\
{}&{}& Gmean\_no\_rej&$23.81$ &$9.96$ &$19.74$ &$80.26$ &--- &---  &--- &$82.81$&$72.81$ &$0.3331$ \\
{}&{} &NI\_no\_rej&$26.03$ &$7.94$ &$20.72$ &$79.28$ &--- &---  &--- &$82.49$&$72.29$ &$0.3359$ \\
{}&{} &Chow's reject&$1.17$  &$13.14$ &$4.69$  &$93.05$ &$20.36$ &$62.33$ &$32.68$&$80.01$ &$75.37$ &$0.2679$ \\
{}&{} &Gmean\_rej&$\bm{0.17}$  &$\bm{0.14}$ &$\bm{0.16}$  &$\bm{99.25}$ &$78.47$ &$91.71$ &$82.36$&$\bm{96.97}$ &$\bm{95.24}$ &$0.1292$ \\
{}&{} &NI\_rej&$10.05$ &$4.35$ &$8.37$ &$88.92$ &$23.50$ &$28.33$ &$24.92$ &$90.32$ &$82.69$ &$\bm{0.3845}$\\\cline{2-12}
\hline 
\multirow{16}{*}{German}&\multirow{8}{0.1in}{$k$\\N\\N} & $k$NN classifier&$9.67$ &$69.03$ &$27.48$ &$72.52$ &--- &---  &--- &$52.75$&$40.24$ &$0.0554$ \\
{}&{} &SMOTE&$49.80$ &$21.39$ &$41.28$ &$58.72$  &--- &--- &--- &$60.30$ &$53.23$ &$0.0746$ \\
{}&{}& Cost-sensitive&$29.57$ &$35.53$ &$31.36$ &$68.64$ &--- &--- &---&$67.32$ &$55.19$ &$0.0888$ \\
{}&{}& Gmean\_no\_rej&$32.92$  &$29.47$ &$31.88$  &$68.12$ &--- &--- &---&$68.67$ &$57.03$ &$0.1017$ \\
{}&{} &NI\_no\_rej &$34.40$ &$28.00$ &$32.48$ &$67.52$ &--- &--- &---&$68.38$ &$57.12$ &$0.1037$ \\
{}&{} &Chow's reject&$\bm{2.17}$ &$35.53$ &$12.18$ &$\bm{81.03}$ &$27.40$ &$55.60$ &$35.86$ &$43.97$ &$30.40$ &$0.0582$\\
{}&{} &Gmean\_rej&$6.20$ &$\bm{6.07}$ &$\bm{6.16}$ &$80.66$ &$68.63$ &$72.53$ &$69.80$ &$\bm{78.48}$ &$\bm{67.65}$ &$0.0797$\\
{}&{} &NI\_rej&$17.30$ &$16.83$ &$17.16$ &$73.89$ &$34.85$ &$33.50$ &$34.45$ &$73.63$ &$63.57$ &$\bm{0.1139}$\\\cline{2-13}
{}&\multirow{8}{0.1in}{B\\a\\y\\e\\s}& Bayes classifier &$\bm{0.00}$ &$100.00$ &$30.03$&$69.97$ &--- &---  &--- &$0.00$&$0.00$ &$0.0000$ \\
{}&{} &SMOTE&$61.84$ &$32.33$ &$52.99$ &$47.01$ &--- &---  &--- &$13.06$ &$37.16$ &$0.0183$ \\
{}&{} & Cost-sensitive&$21.43$ &$39.27$ &$26.78$ &$73.22$ &--- &---  &--- &$69.06$ &$57.64$ &$0.1179$ \\
{}&{}& Gmean\_no\_rej&$32.23$ &$26.73$ &$30.58$ &$69.42$ &--- &---  &--- &$70.23$&$58.96$ &$0.1144$ \\
{}&{} &NI\_no\_rej&$30.80$ &$29.50$ &$30.41$ &$69.59$ &--- &---  &--- &$69.12$&$57.87$ &$0.1189$ \\
{}&{} &Chow's reject&$\bm{0.00}$  &$30.60$ &$9.18$  &$84.62$ &$28.51$ &$69.40$ &$40.78$&$0.00$ &$0.00$ &$0.0567$ \\
{}&{} &Gmean\_rej&$1.77$  &$\bm{2.07}$ &$\bm{1.86}$  &$\bm{89.16}$ &$82.11$ &$88.33$ &$83.98$&$\bm{79.02}$ &$\bm{68.39}$ &$0.0689$ \\
{}&{} &NI\_rej&$15.40$ &$12.23$ &$14.45$ &$76.33$ &$40.20$ &$39.50$ &$39.99$ &$75.88$ &$66.92$ &$\bm{0.1334}$\\
\hline 
\multicolumn{13}{r}{\small\sl (Continued on next page)}\\
\end{tabular}

\end{center}

\end{table*}

\begin{table*}[htbp]\scriptsize
\begin{center}
\begin{tabular}{|l|c|l|c|c|c|c|c|c|c|c|c|c|}
\multicolumn{13}{l}{\small\sl (Continued from previous page)}\\
 \hline
Data set & \multicolumn{2}{c|}{Method} &$E_{N}$(\%) & $E_{P}$(\%)& $E$(\%) & $A$(\%)  &  $Rej_{N}$(\%)  & $Rej_{P}$(\%) & $Rej$(\%) & $G$(\%)&$F$(\%)  & $NI$\\
\hline 
\multirow{16}{*}{Diabetes}&\multirow{8}{0.1in}{$k$\\N\\N} & $k$NN classifier&$13.56$ &$48.33$ &$25.69$ &$74.31$ &--- &---  &--- &$66.72$&$58.29$ &$0.1288$ \\
{}&{} &SMOTE&$42.46$ &$17.96$ &$33.91$ &$66.09$  &--- &--- &--- &$67.45$ &$62.80$ &$0.1320$ \\
{}&{}& Cost-sensitive&$25.00$ &$30.16$ &$26.80$ &$73.20$ &--- &--- &---&$72.33$ &$64.50$ &$0.1495$ \\
{}&{}& Gmean\_no\_rej&$31.64$  &$23.13$ &$28.67$  &$71.33$ &--- &--- &---&$72.37$ &$65.17$ &$0.1538$ \\
{}&{} &NI\_no\_rej &$33.21$ &$21.55$ &$29.14$ &$70.86$ &--- &--- &---&$72.07$ &$65.24$ &$0.1580$ \\
{}&{} &Chow's reject&$\bm{5.16}$ &$26.35$ &$12.56$ &$81.61$ &$24.00$ &$46.48$ &$31.84$ &$68.65$ &$60.00$ &$0.1370$\\
{}&{} &Gmean\_rej&$5.79$ &$\bm{4.32}$ &$\bm{5.28}$ &$\bm{86.66}$ &$64.01$ &$68.46$ &$65.57$ &$\bm{85.55}$ &$\bm{78.96}$ &$0.1381$\\
{}&{} &NI\_rej&$15.28$ &$11.60$ &$14.00$ &$79.25$ &$33.38$ &$33.13$ &$33.29$ &$79.81$ &$73.62$ &$\bm{0.1856}$\\\cline{2-13}
{}&\multirow{8}{0.1in}{B\\a\\y\\e\\s}& Bayes classifier&$1.50$ &$86.38$ &$31.12$ &$68.88$ &--- &---  &--- &$36.19$&$23.17$ &$0.0501$ \\
{}&{} &SMOTE&$53.33$ &$23.66$ &$42.98$ &$57.02$ &--- &---  &--- &$46.01$ &$52.79$ &$0.0861$ \\
{}&{} & Cost-sensitive&$19.16$ &$41.87$ &$27.08$ &$72.92$ &--- &---  &--- &$68.51$ &$59.95$ &$0.1208$ \\
{}&{}& Gmean\_no\_rej&$28.76$ &$23.66$ &$26.98$ &$73.02$ &--- &---  &--- &$73.57$&$66.29$ &$0.1611$ \\
{}&{} &NI\_no\_rej&$37.72$ &$16.96$ &$30.48$ &$69.52$ &--- &---  &--- &$71.37$&$65.56$ &$0.1629$ \\
{}&{} &Chow's reject& $\bm{0.12}$ &$24.25$ &$8.54$  &$84.37$ &$29.56$ &$75.15$ &$45.47$&$10.16$ &$4.25$ &$0.0827$ \\
{}&{} &Gmean\_rej&$1.44$  &$\bm{1.57}$ &$\bm{1.49}$  &$\bm{93.27}$ &$78.97$ &$88.72$ &$82.37$&$\bm{91.72}$ &$\bm{84.75}$ &$0.0971$ \\
{}&{} &NI\_rej&$22.14$ &$8.88$ &$17.51$ &$76.39$ &$26.03$ &$28.93$ &$27.04$ &$78.07$ &$71.68$ &$\bm{0.1893}$\\\cline{2-12}
\hline 
\multirow{16}{*}{Gamma}&\multirow{8}{0.1in}{$k$\\N\\N} & $k$NN classifier&$5.49$ &$35.11$ &$15.90$ &$84.10$ &--- &---  &--- &$78.31$&$74.15$ &$0.3258$ \\
{}&{} &SMOTE&$27.12$ &$15.56$ &$23.06$ &$76.94$  &--- &--- &--- &$77.88$ &$72.23$ &$0.2624$ \\
{}&{}& Cost-sensitive&$11.27$ &$25.35$ &$16.22$ &$83.78$ &--- &--- &---&$81.38$ &$76.40$ &$0.3251$ \\
{}&{}& Gmean\_no\_rej& $17.27$ &$19.21$ &$17.95$  &$82.05$ &--- &--- &---&$81.75$ &$76.00$ &$0.3102$ \\
{}&{} &NI\_no\_rej &$10.13$ &$26.79$ &$15.99$ &$84.01$ &--- &--- &---&$81.11$ &$76.30$ &$0.3275$ \\
{}&{} &Chow's reject&$2.14$ &$22.93$ &$9.45$ &$88.79$ &$11.31$ &$23.87$ &$15.72$ &$82.58$ &$79.83$ &$0.3640$\\
{}&{} &Gmean\_rej&$\bm{1.01}$ &$\bm{3.88}$ &$\bm{2.02}$ &$\bm{95.61}$ &$55.00$ &$52.61$ &$54.16$ &$\bm{94.74}$ &$\bm{93.82}$ &$0.3367$\\
{}&{} &NI\_rej&$3.95$ &$10.84$ &$6.37$ &$91.21$ &$27.28$ &$28.59$ &$27.74$ &$89.57$ &$87.01$ &$\bm{0.3893}$\\\cline{2-13}
{}&\multirow{8}{0.1in}{B\\a\\y\\e\\s}& Bayes classifier&$5.04$ &$47.20$ &$19.87$ &$80.13$ &--- &---  &--- &$70.81$&$65.15$ &$0.2348$ \\
{}&{} &SMOTE&$25.88$ &$22.60$ &$24.73$ &$75.27$ &--- &---  &--- &$74.52$ &$68.70$ &$0.2183$ \\
{}&{} & Cost-sensitive&$13.38$ &$31.31$ &$19.68$ &$80.32$ &--- &---  &--- &$77.13$ &$71.05$ &$0.2459$ \\
{}&{}& Gmean\_no\_rej&$21.82$ &$22.45$ &$22.04$ &$77.96$ &--- &---  &--- &$77.83$&$71.23$ &$0.2336$ \\
{}&{} &NI\_no\_rej&$12.02$ &$33.29$ &$19.50$ &$80.50$ &--- &---  &--- &$76.59$&$70.62$ &$0.2461$ \\
{}&{} &Chow's reject&$0.81$  &$25.51$ &$9.49$  &$87.16$ &$18.18$ &$40.67$ &$26.09$&$75.13$ &$71.48$ &$0.2634$ \\
{}&{} &Gmean\_rej& $\bm{0.05}$ &$\bm{0.11}$ &$\bm{0.07}$  &$\bm{99.30}$ &$91.65$ &$86.48$ &$89.83$&$\bm{99.22}$ &$\bm{99.25}$ &$0.1070$ \\
{}&{} &NI\_rej&$3.87$ &$10.44$ &$6.18$ &$90.13$ &$36.09$ &$40.40$ &$37.60$ &$88.00$ &$84.85$ &$\bm{0.3066}$\\
\hline 

\end{tabular}

\end{center}
\end{table*}

\begin{table*}[htbp]\scriptsize
\begin{center}
\caption{The \textit{``Equivalent''} Costs and the Optimal Rejection Thresholds for Binary Class Data Sets}
 \label{result2}
\subtable[$k$NN Classifier  Based]{
\setlength{\extrarowheight}{0.6ex}
\begin{tabular}{|l|c|c|c|c|c|}
\hline
Data set &$\alpha^{\ast}_{N}\big(\bar{\lambda}_{FP}\big)$ &$T^{\ast}_{rN}$ &$T^{\ast}_{rP}$ &$\bar{\lambda}_{RN}$ &$\bar{\lambda}_{RP}$ \\
\hline
Ism&$0.2312(0.0408)$&$0.0743(0.0085)$ &$0.7643(0.0296)$ &$0.0272$&$0.6616$\\
Nursery&$0.3482(0.0328)$ &$0.1215(0.0565)$ &$0.7125(0.0403)$ &$0.0288$ &$0.7914$ \\
Letter&$0.3802(0.0321)$ &$0.1284(0.0343)$ &$0.6140(0.0517)$ &$0.0760$ &$0.4838$ \\
Rooftop&$0.1372(0.0302)$ &$0.0705(0.0610)$ &$0.7733(0.0404)$ &$0.0544$ &$0.2823$ \\
Pendigits&$0.4714(0.0771)$ &$0.1745(0.0851)$ &$0.3890(0.0555)$ &$0.1710$ &$0.1913$ \\
Optdigits&$0.4061(0.0667)$ &$0.1487(0.0372)$ &$0.5913(0.0486)$ &$0.0964$ &$0.4481$ \\
Vehicle&$0.1972(0.0769)$ &$0.1101(0.0297)$ &$0.6266(0.0719)$ &$0.1045$ &$0.1556$ \\
Yeast&$0.3610(0.1333)$ &$0.1245(0.0335)$ &$0.5608(0.0536)$ &$0.0937$ &$0.3414$ \\
Phoneme&$0.4651(0.0835)$ &$0.1319(0.0180)$ &$0.4543(0.0409)$ &$0.1066$ &$0.2985$ \\
German&$0.3848(0.0785)$ &$0.1915(0.0336)$ &$0.6021(0.0368)$ &$0.1542$ &$0.3489$ \\
Diabetes&$0.3725(0.0796)$ &$0.1725(0.0455)$ &$0.5284(0.0672)$ &$0.1585$ &$0.2398$ \\
Gamma&$0.5682(0.0207)$ &$0.1663(0.0225)$ &$0.4188(0.0319)$ &$0.1376$ &$0.3103$ \\

\hline

\end{tabular}
}\\
\subtable[Bayes Classifier Based]{
\setlength{\extrarowheight}{0.6ex}
\begin{tabular}{|l|c|c|c|c|c|}
\hline
Data set &$\alpha^{\ast}_{N}\big(\bar{\lambda}_{FP}\big)$ &$T^{\ast}_{rN}$ &$T^{\ast}_{rP}$ &$\bar{\lambda}_{RN}$ &$\bar{\lambda}_{RP}$ \\
\hline
Ism&$0.1420(0.0230)$ &$0.0222(0.0168)$ &$0.8560(0.0212)$ &$0.0041$ &$0.8198$\\
Nursery&$0.1052(0.0117)$ &$0.0593(0.0076)$ &$0.8845(0.0090)$ &$0.0237$ &$0.6242$\\
Letter&$0.1155(0.0066)$ &$0.0687(0.0163)$ &$0.8772(0.0128)$ &$0.0273$ &$0.6302$\\
Rooftop&$0.0786(0.0299)$ &$0.0242(0.0080)$ &$0.8363(0.0242)$ &$0.0170$ &$0.3147$\\
Pendigits&$0.4283(0.0472)$ &$0.1521(0.0488)$ &$0.6170(0.0394)$ &$0.0782$ &$0.5640$\\
Optdigits&$0.1883(0.0087)$ &$0.1339(0.0047)$ &$0.8274(0.0078)$ &$0.0581$ &$0.6240$\\
Vehicle&$0.2490(0.0262)$ &$0.1301(0.0149)$ &$0.5369(0.0384)$ &$0.1287$ &$0.1395$\\
Yeast &$0.4469(0.0734)$ &$0.2668(0.0147)$ &$0.6413(0.0137)$ &$0.2093$&$0.4247$\\
Phoneme&$0.3364(0.0374)$ &$0.1723(0.0266)$ &$0.5511(0.0363)$ &$0.1641$ &$0.2115$\\
German &$0.4265(0.0137)$ &$0.2802(0.0081)$ &$0.6866(0.0085)$ &$0.1736$&$0.5541$\\
Diabetes&$0.3808(0.0460)$ &$0.2461(0.0174)$ &$0.6638(0.0327)$ &$0.2279$ &$0.3020$\\
Gamma&$0.5859(0.0483)$ &$0.1723(0.0212)$ &$0.4660(0.0272)$ &$0.1243$ &$0.4028$\\
\hline
\end{tabular}
}

\footnotesize
\hspace{-1.4in}Optimal values are listed as  mean(standard deviation). 

\hspace{-0.5in} (a) Derived based on $k$NN classifier. (b) Derived based on Bayes classifier. 
\end{center}
\end{table*}

\begin{figure*}[ht]
\centering
\mbox{
\hspace{-0.3in}
\subfigure[For non-abstaining classification]{\includegraphics[scale=0.50]{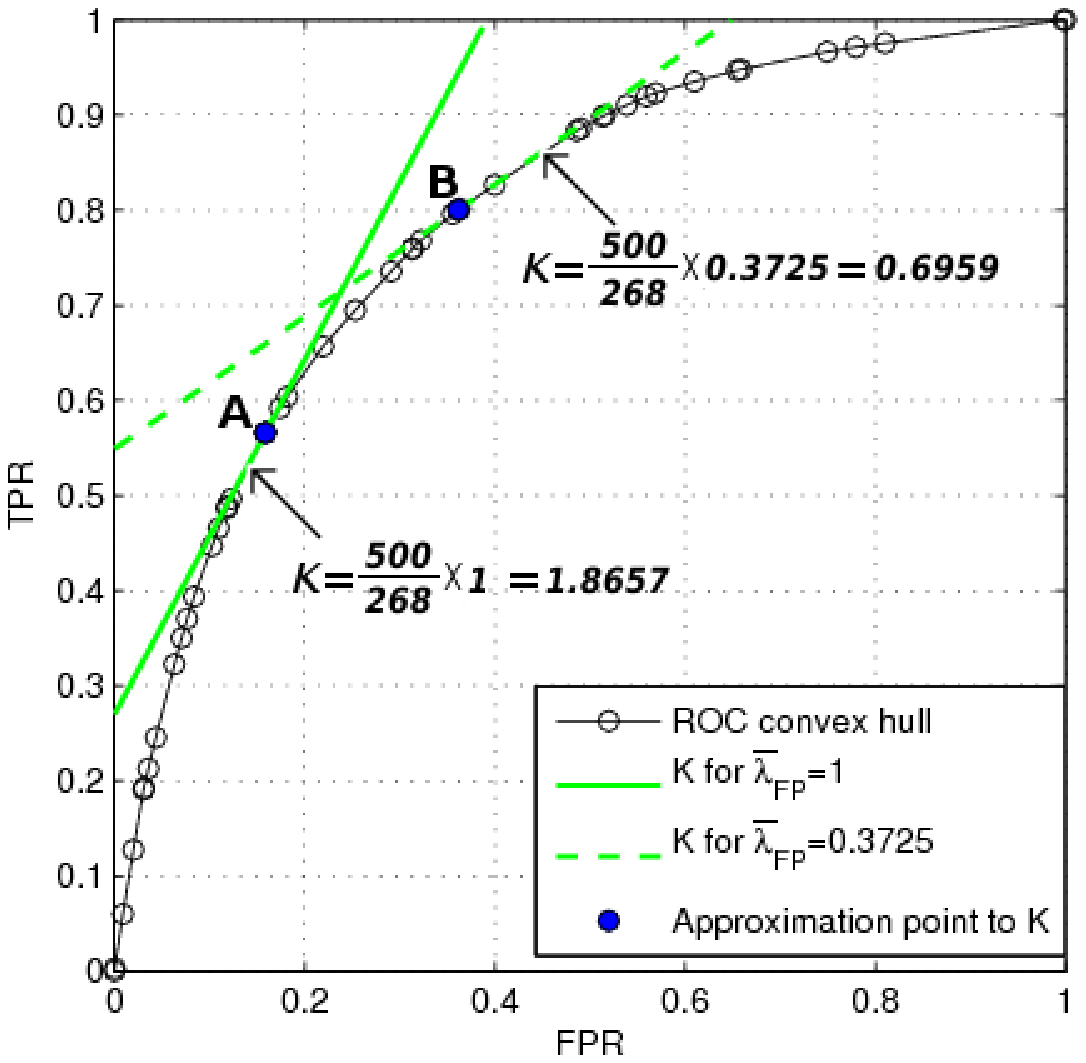}} 
\hspace{-0.3in}
\subfigure[For abstaining classification]{\includegraphics[scale=0.50]{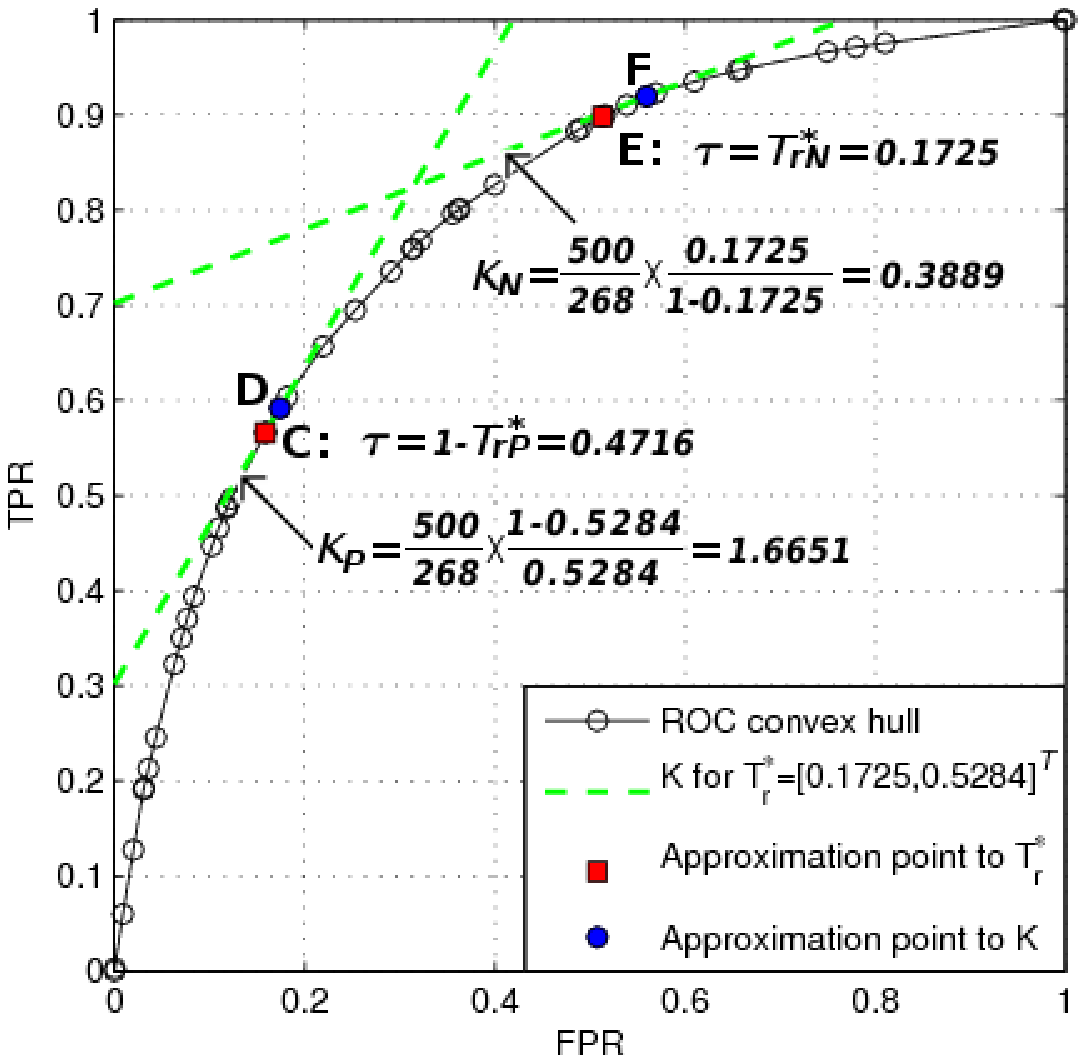}} 
}
\caption{Results on ROCCH of $k$NN for Diabetes. (a) For non-abstaining classification. (b) For abstaining classification.} \label{figROC2}
\end{figure*}

\subsection{Binary Class Tasks}
The  results on the binary class data sets are shown in Table {\ref{result1}}. Both conventional classifiers have high  accuracies and low error rates of the negative class, but the error rates of the positive class are high. \textit{SMOTE} is an  effective method  with low error rate of the positive class. However, it does not have the ability to reject instances. 
\textit{Cost-sensitive learning} performs well under the current cost settings, but its accuracy is the lowest when the class distribution differs greatly.
On \textit{Nursery} and \textit{Letter}, the error rate of the positive class is zero with \textit{Cost-sensitive learning} at the price of high error rate of the negative class.
Besides, \textit{Gmean\_no\_rej} and \textit{NI\_no\_rej} perform well on balancing the classification of two classes.
When a reject option is added, the error rate may be reduced and the accuracy may be increased. But it is difficult to decide the rejection costs and the rejection thresholds for lack of information about the rejections. Regarding to  \textit{Chow's reject},  it is  usually wasteful to reject  lots of instances from the positive  class with arbitrary settings on the rejection thresholds. 
On most  data sets, \textit{Gmean\_rej} achieves the highest accuracy and the lowest error rate of the positive class,  at the price of considerably high reject rate. However,  the accuracy of \textit{Gmean\_rej} is lower than  the conventional classifications on \textit{Ism} and \textit{Rooftop}. One explanation  is that the goal of the G-mean based methods is to maximize the geometric mean of the accuracy within each class rather than the total accuracy. 
Compared with \textit{Gmean\_rej}  and \textit{Chow's reject}, our  \textit{NI\_rej} performs    best on the whole with low error rate of the positive class, high accuracy, a certain amount of reject rate, high G-mean, high F-measure and the highest NI.

Table {\ref{result2}} lists the values of  the optimal  weight $\alpha^{\ast}_{N}$ and rejection thresholds $\bm{T^{\ast}_{r}}$.
The last two columns represent the ``\textit{equivalent}'' rejection costs  computed with the mean values of  $\alpha^{\ast}_{N}$ and $\bm{T^{\ast}_{r}}$. Moreover,  these values are purely determined by the data sets besides the conventional classification algorithms. They can be adopted as ``\textit{objective}'' references while the cost information is unknown.
In addition, the \textit{``equivalent''} costs   of these data sets are consistent with
 human assumption, which also reflects the effectiveness of our NI based strategy.

\begin{table}[ht]\scriptsize
\begin{center}
\caption{Some ROCCH Vertices of $k$NN  for Diabetes} \label{ROC}
\begin{tabular}{|c|c|c|c|} \hline
Index& ROC Convex Hull &Slope&Threshold\\
(Point Label)&Vertices (FPR, TPR)&$\widehat{K}$&$\tau$\\
 \hline
  17 \hspace{0.2in}  &(0.1226,    0.4965)& 2.1593    &0.5106\\
  \hspace{0.03in} $\bm{18}$(\bf{A}, \bf{C}\textnormal{)} &$\bm{(0.1585,    0.5656)}$& $\bm{1.9255}$    &$\bm{0.4514}$\\
   19 \hspace{0.2in} &(0.1588,    0.5662)& 1.8502  &0.4508\\
   $\bm{20}$(\bf{D}\textnormal{)}\hspace{0.07in} &$\bm{(0.1739,    0.5915)}$& $\bm{1.6778}$   &$\bm{0.4325}$\\
   21 \hspace{0.19in} &(0.1816,    0.6038)& 1.5962   &0.4211\\
  \dots &\dots&\dots&\dots\\
   28 \hspace{0.19in} &(0.3554,    0.7958)& 0.8326    &0.2617\\
   $\bm{29}$(\bf{B}\textnormal{)}\hspace{0.07in} &$\bm{(0.3615,    0.8002)}$& $\bm{0.7357}$    &$\bm{0.2586}$\\
   30 \hspace{0.19in} &(0.3634,    0.8016)& 0.6921    &0.2578\\
   31 \hspace{0.19in} &(0.3997,    0.8265)& 0.6881    &0.2334\\
   32 \hspace{0.19in} &(0.4860,    0.8832)& 0.6554    &0.1781\\
   33 \hspace{0.19in} &(0.4901,    0.8858)& 0.6543   &0.1769\\
   $\bm{34}$(\bf{E}\textnormal{)}\hspace{0.07in} &$\bm{(0.5129,    0.8983)}$& $\bm{0.5475}$    &$\bm{0.1710}$\\
   35 \hspace{0.19in} &(0.5158,    0.8998)& 0.5113    &0.1701\\
   36 \hspace{0.19in} &(0.5389,    0.9104)& 0.4605   &0.1632\\
   $\bm{37}$(\bf{F}\textnormal{)}\hspace{0.07in} &$\bm{(0.5596,    0.9196)}$&$\bm{0.4434}$    &$\bm{0.1521}$\\
   38 \hspace{0.19in} &(0.5690,    0.9229)& 0.3558   &0.1345\\
\hline
\end{tabular}
\end{center}
\footnotesize
\hspace{0.25in}(A$\sim$F:  point labels shown in Fig. \ref{figROC2} are bolded)
\end{table}

Fig. \ref{figROC2} shows the {\it ROC convex hull} (ROCCH) of \textit{$k$NN} for \textit{Diabetes} generated from $90$ validation sets by threshold averaging \cite{Fawcett2006}. We just list some  of the vertices  in Table \ref{ROC}. And we use them to approximatively locate the parameters \cite{Tortorella2004}. In Fig. \ref{figROC2}a,   we use equal and ``{\it equivalent}'' misclassification costs to compute the slopes, respectively. 
The slopes computed with costs are the same as those computed with rejection thresholds on ROCCH, so we only plot the latter in Fig. \ref{figROC2}b.
Point \textit{B} in Fig. \ref{figROC2}a lies between \textit{D} and \textit{F}.
Due to approximation, points \textit{D} and \textit{F}  that the slopes find are not cohere with the optimal threshold points \textit{C} and \textit{E}.
 In addition, the  parameters can be adjusted with the graphical interpretations  in Fig. \ref{fig2} under the property P3 by the users.

\begin{table*}[htbp]\scriptsize
\begin{center}
\caption{ Results on Cardiotocography and Thyroid} \label{result_cardio}
\subtable[]{
\begin{tabular}{|c|c|l|p{0.77cm}<{\centering}|p{0.77cm}<{\centering}|p{0.74cm}<{\centering}|p{0.63cm}<{\centering}|p{0.72cm}<{\centering}|p{0.955cm}<{\centering}|p{0.955cm}<{\centering}|p{0.955cm}<{\centering}|p{0.83cm}<{\centering}|p{0.73cm}<{\centering}|p{0.87cm}<{\centering}|}
 \hline
Data set & \multicolumn{2}{c|}{Method} &$E_{1}$(\%) & $E_{2}$(\%)& $E_{3}$(\%)&$E$(\%)&  $A$(\%) & $Rej_{1}$(\%) & $Rej_{2}$(\%)& $Rej_{3}$(\%)& $Rej$(\%) &$G$(\%) & $NI$\\
\hline 
&\multirow{8}{0.19cm}{$k$\\N\\N} & $k$NN classifier&$2.39$ &$39.49$ &$28.67$ &$9.71$ &$90.29$ &--- & ---& ---& ---&$74.88$ &$0.4798$ \\ 
&{} &SMOTE&$6.71$ &$23.35$ &$21.26$ &$10.22$ &$89.78$ &--- & ---& ---& ---&$82.37$ &$0.5346$ \\ 
&{} &Cost-sensitive&$100.00$ &$100.00$ &$\bm{0.00}$ &$91.72$ &$8.28$ &--- & ---& ---& ---&$0.00$ &$0.0000$ \\ 
&{} &Gmean\_no\_rej&$13.20$ &$16.33$ &$15.61$ &$13.83$ &$86.17$ &--- & ---& ---& ---&$84.86$ &$0.5105$ \\ 
&{} &NI\_no\_rej&$8.59$ &$19.93$ &$19.95$ &$11.10$ &$88.90$ &--- & ---& ---& ---&$83.56$ &$0.5224$ \\ 
&{} &Chow's reject&$0.53$ &$19.66$ &$15.47$ &$4.42$ &$94.96$ &$5.70$ &$40.54$ &$27.28$ &$12.32$ &$80.56$ &$0.4566$ \\ 
&{}&Gmean\_rej&$\bm{0.39}$ &$\bm{4.88}$ &$5.09$ &$\bm{1.40}$ &$\bm{98.10}$ &$19.23$ &$68.77$ &$31.41$ &$27.11$ &$\bm{91.05}$ &$0.4724$ \\ 
Cardioto-&{}&NI\_rej&$4.44$ &$9.46$ &$15.96$ &$6.09$ &$92.94$ &$13.24$ &$25.08$ &$6.34$ &$14.31$ &$88.04$ &$\bm{0.5574}$ \\    \cline{2-14}
cography&\multirow{8}{0.19cm}{B\\a\\y\\e\\s}& Bayes classifier&$0.06$ &$88.51$ &$76.89$ &$18.69$ &$81.31$ &--- & ---& ---& ---&$29.44$ &$0.1579$ \\ 
&{} &SMOTE&$7.03$ &$41.79$ &$56.52$ &$15.96$ &$84.04$ &--- & ---& ---& ---&$59.41$ &$0.3378$ \\ 
&{} & Cost-sensitive&$100.00$ &$100.00$ &$\bm{0.00}$ &$91.72$ &$8.28$ &--- & ---& ---& ---&$0.00$ &$0.0000$ \\ 
&{} &Gmean\_no\_rej&$17.28$ &$15.81$ &$22.01$ &$17.47$ &$82.53$ &--- & ---& ---& ---&$81.49$ &$0.4383$ \\ 
&{} &NI\_no\_rej&$12.13$ &$22.10$ &$27.06$ &$14.75$ &$85.25$ &--- & ---& ---& ---&$79.13$ &$0.4401$ \\ 
&{} &Chow's reject&$\bm{0.00}$ &$40.40$ &$19.63$ &$7.23$ &$91.21$ &$4.87$ &$59.60$ &$69.11$ &$17.78$ &$0.00$ &$0.1158$ \\ 
&{} &Gmean\_rej&$0.08$ &$\bm{0.88}$ &$6.03$ &$\bm{0.69}$ &$\bm{98.16}$ &$60.57$ &$94.03$ &$53.30$ &$64.61$ &$\bm{89.21}$ &$0.2510$ \\ 
&{} &NI\_rej&$5.59$ &$6.92$ &$30.51$ &$7.84$ &$90.59$ &$17.43$ &$33.81$ &$8.18$ &$18.94$ &$81.99$ &$\bm{0.4666}$ \\ 
\hline 
\multirow{16}{*}{Thyroid}&\multirow{8}{0.19cm}{$k$\\N\\N} & $k$NN classifier&$0.08$ &$94.65$ &$52.17$ &$6.12$ &$93.88$ &--- & ---& ---& ---&$29.11$ &$0.1726$ \\ 
{}&{} &SMOTE&$2.86$ &$79.26$ &$39.69$ &$7.62$ &$92.38$ &--- & ---& ---& ---&$47.84$ &$0.2245$ \\ 
{}&{} &Cost-sensitive&$100.00$ &$100.00$ &$\bm{0.00}$ &$97.69$ &$2.31$ &--- & ---& ---& ---&$0.00$ &$0.0000$ \\ 
{}&{} &Gmean\_no\_rej&$25.69$ &$41.03$ &$22.54$ &$26.41$ &$73.59$ &--- & ---& ---& ---&$\bm{69.65}$ &$0.2233$ \\ 
{}&{} &NI\_no\_rej&$5.85$ &$70.92$ &$32.28$ &$9.78$ &$90.22$ &--- & ---& ---& ---&$56.64$ &$0.2347$ \\ 
{}&{} &Chow's reject&$\bm{0.00}$ &$85.33$ &$30.37$ &$\bm{5.06}$ &$\bm{94.81}$ &$1.05$ &$13.47$ &$35.44$ &$2.47$ &$16.18$ &$0.1277$ \\ 
{}&{}&Gmean\_rej&$9.28$ &$\bm{36.47}$ &$21.10$ &$10.94$ &$86.61$ &$19.16$ &$28.02$ &$10.61$ &$19.42$ &$68.91$ &$0.2413$ \\ 
{}&{}&NI\_rej&$4.68$ &$43.43$ &$22.80$ &$7.08$ &$91.55$ &$16.91$ &$29.39$ &$12.75$ &$17.45$ &$63.93$ &$\bm{0.2555}$ \\    \cline{2-14}
{}&\multirow{8}{0.19cm}{B\\a\\y\\e\\s}& Bayes classifier&$0.06$ &$99.24$ &$87.04$ &$7.14$ &$92.86$ &--- & ---& ---& ---&$7.24$ &$0.0398$ \\ 
{}&{} &SMOTE&$0.89$ &$95.49$ &$78.46$ &$7.51$ &$92.49$ &--- & ---& ---& ---&$18.74$ &$0.0708$ \\ 
{}&{} & Cost-sensitive&$100.00$ &$100.00$ &$\bm{0.00}$ &$97.69$ &$2.31$ &--- & ---& ---& ---&$0.00$ &$0.0000$ \\ 
{}&{} &Gmean\_no\_rej&$35.04$ &$40.83$ &$25.29$ &$35.11$ &$64.89$ &--- & ---& ---& ---&$65.67$ &$0.1885$ \\ 
{}&{} &NI\_no\_rej&$14.40$ &$63.28$ &$34.43$ &$17.36$&$82.64$ &--- & ---& ---& ---&$58.12$ &$0.1921$ \\ 
{}&{} &Chow's reject&$\bm{0.03}$ &$97.29$ &$71.09$ &$6.64$ &$\bm{93.30}$ &$0.30$ &$1.90$ &$20.35$ &$0.84$ &$8.40$ &$0.0301$ \\ 
{}&{} &Gmean\_rej&$2.80$ &$\bm{4.62}$ &$11.29$ &$\bm{3.09}$ &$86.06$ &$77.82$ &$85.94$ &$43.99$ &$77.45$ &$\bm{76.66}$ &$0.1401$ \\ 
{}&{} &NI\_rej&$8.67$ &$31.50$ &$29.03$ &$10.31$ &$85.80$ &$29.45$ &$42.05$ &$11.42$ &$29.68$ &$60.96$ &$\bm{0.1978}$ \\ 
\hline

\end{tabular}
}

\qquad
\subtable[]{
\begin{tabular}{|c|c|c|c|c|}
 \hline
&\multirow{4}{*}{$k$NN}&$\alpha^{\ast}_{1}$ &$\alpha^{\ast}_{2}$ &$\alpha^{\ast}_{3}$ \\
 \cline{3-5}
&{} & $0.2276(0.0592)$  &  $0.7138(0.1076)$  &$1$\\\Xcline{3-5}{0.5pt}
\cline{3-5}
&{}&$T^{\ast}_{r1}$ &$T^{\ast}_{r2}$ &$T^{\ast}_{r3}$ \\
 \cline{3-5}
Cardioto-&{}& $0.0953(0.0391)$  &  $0.6001(0.1053)$ & $0.7122(0.0888)$ \\
\cline{2-5}
cography&\multirow{4}{*}{Bayes}&$\alpha^{\ast}_{1}$ &$\alpha^{\ast}_{2}$ &$\alpha^{\ast}_{3}$  \\
 \cline{3-5}
&{} & $0.1332(0.0265)$  &  $0.5071(0.0697)$  &$1$\\\Xcline{3-5}{0.5pt}
\cline{3-5}
&{}&$T^{\ast}_{r1}$ &$T^{\ast}_{r2}$ &$T^{\ast}_{r3}$ \\
 \cline{3-5}
&{}& $0.1530(0.0220)$  &  $0.7366(0.0422)$ & $0.8249(0.0362)$ \\
\hline
\hline
\multirow{8}{*}{Thyroid}&\multirow{4}{*}{$k$NN}&$\alpha^{\ast}_{1}$ &$\alpha^{\ast}_{2}$ &$\alpha^{\ast}_{3}$ \\
 \cline{3-5}
{}&{}& $0.1505(0.0395)$  &  $0.7065(0.1291)$  &$1$\\\Xcline{3-5}{0.5pt}
\cline{3-5}
{}&{}&$T^{\ast}_{r1}$ &$T^{\ast}_{r2}$ &$T^{\ast}_{r3}$ \\
 \cline{3-5}
{}&{}& $0.0764(0.0269)$  &  $0.8058(0.0486)$ & $0.8332(0.0450)$ \\
\cline{2-5}
{}&\multirow{4}{*}{Bayes}&$\alpha^{\ast}_{1}$ &$\alpha^{\ast}_{2}$ &$\alpha^{\ast}_{3}$  \\
 \cline{3-5}
{}&{}& $0.0367(0.0069)$  &  $0.4037(0.0751)$  &$1$\\\Xcline{3-5}{0.5pt}
\cline{3-5}
{}&{}&$T^{\ast}_{r1}$ &$T^{\ast}_{r2}$ &$T^{\ast}_{r3}$ \\
 \cline{3-5}
{}&{}& $0.0726(0.0171)$  &  $0.8996(0.0336)$ & $0.9553(0.0219)$ \\
\hline

\end{tabular}
}

\end{center}
\footnotesize
(a) Evaluation of the methods, ``--'': not available, the best performance in each cell is bolded.

(b) Optimal weights and rejection thresholds are listed as mean(standard deviation). 
\end{table*}

\begin{table*}[htbp]\scriptsize
\begin{center}
\caption{Results on Car} \label{result_car}
\subtable[]{
\begin{tabular}{|c|c|l|p{0.77cm}<{\centering}|p{0.77cm}<{\centering}|p{0.77cm}<{\centering}|p{0.63cm}<{\centering}|p{0.72cm}<{\centering}|p{0.955cm}<{\centering}|p{0.955cm}<{\centering}|p{0.955cm}<{\centering}|p{0.83cm}<{\centering}|p{0.73cm}<{\centering}|p{0.87cm}<{\centering}|}
 \hline
Data set &\multicolumn{2}{c|}{ Method} &$E_{1}$(\%) & $E_{3}$(\%)& $E_{4}$(\%)&$E$(\%)&  $A$(\%) & $Rej_{1}$(\%) & $Rej_{3}$(\%)& $Rej_{4}$(\%)& $Rej$(\%) &$G$(\%) & $NI$\\
\hline 
\multirow{16}{*}{Car}&\multirow{8}{0.19cm}{$k$\\N\\N} & $k$NN classifier&$0.50$ &$35.76$ &$62.03$ &$7.82$ &$92.18$ &--- & ---& ---& ---&$66.65$ &$0.6357$ \\ 
{} &{} &SMOTE&$7.13$ &$29.08$ &$52.41$ &$9.25$ &$90.75$ &--- & ---& ---& ---&$73.23$ &$0.6554$ \\
{} &{} &Cost-sensitive&$100.00$ &$\bm{0.00}$ &$100.00$ &$96.24$ &$3.76$ &--- & ---& ---& ---&$0.00$ &$0.0000$ \\
{} &{} &Gmean\_no\_rej&$6.13$ &$6.16$ &$7.54$ &$8.55$ &$91.45$ &--- & ---& ---& ---&$90.62$ &$0.7091$ \\
{} &{} &NI\_no\_rej &$3.31$ &$13.05$ &$24.93$ &$6.17$ &$93.83$ &--- & ---& ---& ---&$86.28$ &$0.7356$ \\
{} &{} &Chow's reject&$\bm{0.00}$ &$1.45$ &$\bm{1.45}$ &$\bm{0.23}$ &$\bm{99.70}$ &$5.70$ &$78.47$ &$97.10$  &$24.37$  &$50.00$ &$0.4574$ \\ 
{} &{}&Gmean\_rej&$0.05$ &$4.33$ &$4.06$ &$0.94$ &$98.84$ &$11.44$ &$23.64$ &$42.90$  &$21.50$  &$\bm{95.47}$ &$0.6265$ \\ 
{} &{}&NI\_rej&$1.94$ &$9.70$ &$20.87$ &$4.35$ &$95.34$ &$6.66$ &$0.30$ &$2.46$  &$7.00$  &$88.68$ &$\bm{0.7635}$ \\      \cline{2-14}
{} &\multirow{8}{0.19cm}{B\\a\\y\\e\\s}& Bayes classifier &$\bm{0.00}$ &$100.00$ &$100.00$ &$29.98$ &$70.02$ &--- & ---& ---& ---&$0.00$ &$0.0000$ \\
{} &{} &SMOTE &$27.51$ &$100.00$ &$100.00$ &$34.09$ &$65.91$ &--- & ---& ---& ---&$0.00$ &$0.1733$ \\
{} &{} & Cost-sensitive &$100.00$ &$\bm{0.00}$ &$100.00$ &$96.24$ &$3.76$ &--- & ---& ---& ---&$0.00$ &$0.0000$ \\
{} &{} &Gmean\_no\_rej &$26.48$ &$17.07$ &$12.75$ &$24.06$ &$75.94$ &--- & ---& ---& ---&$80.44$ &$0.4302$ \\
{} &{} &NI\_no\_rej &$26.05$ &$15.40$ &$21.88$ &$23.25$ &$76.75$ &--- & ---& ---& ---&$79.38$ &$0.4374$ \\
{} &{} &Chow's reject&$\bm{0.00}$ &$\bm{0.00}$ &$\bm{0.00}$ &$\bm{1.04}$ &$\bm{98.07}$ &$24.73$ &$100.00$ &$100.00$  &$46.25$  &$0.00$ &$0.1828$ \\ 
{} &{} &Gmean\_rej&$10.35$ &$7.72$ &$10.14$ &$10.47$ &$87.33$ &$24.02$ &$25.98$ &$44.64$  &$34.06$  &$\bm{81.33}$ &$0.3486$ \\ 
{} &{} &NI\_rej&$18.76$ &$14.59$ &$14.64$ &$17.76$ &$80.44$ &$10.10$ &$0.14$ &$8.12$  &$11.86$  &$79.64$ &$\bm{0.4388}$ \\ 
\hline

\end{tabular}
}
\qquad\\
\subtable[]{
\begin{tabular}{|l|c|c|c|c|}
 \hline
\multirow{4}{*}{$k$NN}&$\alpha^{\ast}_{1}$ &$\alpha^{\ast}_{2}$ &$\alpha^{\ast}_{3}$ &$\alpha^{\ast}_{4}$  \\
 \cline{2-5}
{} & $0.3029(0.0713)$  &  $0.5127(0.1216)$ & $1.0183(0.1751)$ &$1$\\\Xcline{2-5}{0.5pt}
\cline{2-5}
{}&$T^{\ast}_{r1}$ &$T^{\ast}_{r2}$ &$T^{\ast}_{r3}$ &$T^{\ast}_{r4}$ \\
 \cline{2-5}
{}& $0.2509(0.0514)$  &  $0.5751(0.0455)$ & $0.7885(0.0604)$ &$0.8027(0.0394)$ \\
\hline
\hline
\multirow{4}{*}{Bayes}&$\alpha^{\ast}_{1}$ &$\alpha^{\ast}_{2}$ &$\alpha^{\ast}_{3}$ &$\alpha^{\ast}_{4}$  \\
 \cline{2-5}
{} & $0.0997(0.0072)$  &  $0.2825(0.0171)$ & $0.9958(0.0984)$ &$1$\\\Xcline{2-5}{0.5pt}
\cline{2-5}
{}&$T^{\ast}_{r1}$ &$T^{\ast}_{r2}$ &$T^{\ast}_{r3}$ &$T^{\ast}_{r4}$ \\
 \cline{2-5}
{}& $0.3083(0.0379)$  &  $0.7365(0.0340)$ & $0.9209(0.0104)$ &$0.9245(0.0115)$ \\
\hline

\end{tabular}
}
\end{center}
\footnotesize
(a) Evaluation of the methods, ``--'': not available, the best performance in each cell is bolded.

(b) Optimal weights and rejection thresholds are listed as mean(standard deviation).
\end{table*}

\begin{table*}[ht]\scriptsize
\begin{center}
\caption{Results on Pageblock} \label{result_pageblock}
\subfigure[]{
\begin{tabular}{|p{1.1cm}<{\centering}|c|l|p{0.77cm}<{\centering}|p{0.70cm}<{\centering}|p{0.77cm}<{\centering}|p{0.63cm}<{\centering}|p{0.72cm}<{\centering}|p{0.955cm}<{\centering}|p{0.955cm}<{\centering}|p{0.955cm}<{\centering}|p{0.83cm}<{\centering}|p{0.73cm}<{\centering}|p{0.87cm}<{\centering}|}
 \hline
Data set & \multicolumn{2}{c|}{Method} &$E_{1}$(\%) & $E_{3}$(\%)& $E_{5}$(\%)&$E$(\%)&  $A$(\%) & $Rej_{1}$(\%) & $Rej_{3}$(\%)& $Rej_{5}$(\%)& $Rej$(\%) &$G$(\%) & $NI$\\
\hline 
\multirow{16}{*}{Pageblock}&\multirow{8}{0.2cm}{$k$\\N\\N} & $k$NN (k=5)&$1.08$ &$42.30$ &$51.03$ &$4.12$ &$95.88$ &--- & ---& ---& ---&$68.12$ &$0.5466$ \\ 
{}&{} &SMOTE&$2.46$ &$32.65$ &$40.10$ &$4.78$ &$95.22$ &--- & ---& ---& ---&$74.10$ &$0.5666$ \\ 
{}&{} &Cost-sensitive&$100.00$ &$\bm{0.00}$ &$100.00$ &$99.49$ &$0.51$ &--- & ---& ---& ---&$0.00$ &$0.0000$ \\ 
{}&{} &Gmean\_no\_rej&$6.04$ &$9.26$ &$24.48$ &$7.08$ &$92.92$ &--- & ---& ---& ---&$84.74$ &$0.5899$ \\ 
{}&{} &NI\_no\_rej&$3.73$ &$15.30$ &$26.66$ &$5.37$ &$94.63$ &--- & ---& ---& ---&$82.17$ &$0.6019$ \\ 
{}&{} &Chow's reject&$\bm{0.49}$ &$22.74$ &$33.56$ &$\bm{2.50}$ &$\bm{97.40}$ &$2.06$ &$36.37$ &$35.13$ &$4.24$ &$71.33$ &$0.5022$ \\ 
{}&{}&Gmean\_rej&$2.13$ &$4.30$ &$\bm{19.84}$ &$3.18$ &$96.66$ &$4.00$ &$12.37$ &$19.79$ &$5.32$ &$\bm{86.94}$ &$0.6028$ \\ 
{}&{}&NI\_rej&$2.57$ &$11.52$ &$21.53$ &$3.75$ &$96.10$ &$3.23$ &$5.89$ &$12.26$ &$3.99$ &$85.52$ &$\bm{0.6179}$ \\     \cline{2-14}
{}&\multirow{8}{0.2cm}{B\\a\\y\\e\\s}& Bayes classifier&$0.78$ &$56.30$ &$62.86$ &$5.15$ &$94.85$ &--- & ---& ---& ---&$57.89$ &$0.4559$ \\
{}&{} &SMOTE&$2.19$ &$51.61$ &$41.73$ &$5.53$ &$94.47$ &--- & ---& ---& ---&$66.09$ &$0.4982$ \\
{}&{} & Cost-sensitive&$100.00$ &$\bm{0.00}$ &$100.00$ &$99.49$ &$0.51$ &--- & ---& ---& ---&$0.00$ &$0.0000$ \\
{}&{} &Gmean\_no\_rej&$14.45$ &$17.06$ &$16.23$ &$14.77$ &$85.23$ &--- & ---& ---& ---&$84.66$ &$0.4900$ \\
{}&{} &NI\_no\_rej&$13.71$ &$17.33$ &$15.47$ &$14.22$ &$85.78$ &--- & ---& ---& ---&$82.33$ &$0.5055$ \\
{}&{} &Chow's reject&$\bm{0.17}$ &$47.85$ &$37.39$ &$2.92$ &$\bm{96.91}$ &$2.04$ &$9.93$ &$41.91$ &$5.59$ &$50.21$ &$0.3362$ \\ 
{}&{} &Gmean\_rej&$1.26$ &$7.26$ &$\bm{5.59}$ &$\bm{1.66}$ &$96.17$ &$50.29$ &$25.48$ &$60.13$ &$49.46$ &$\bm{87.78}$ &$0.4714$ \\ 
{}&{} &NI\_rej&$2.41$ &$11.04$ &$14.71$ &$3.30$ &$96.25$ &$11.27$ &$19.26$ &$23.13$ &$12.28$ &$87.15$ &$\bm{0.5749}$ \\ 
\hline 

\end{tabular}
}

\qquad  \\
\subfigure[] {
\begin{tabular}{|l|c|c|c|c|c|}
 \hline
\multirow{4}{*}{$k$NN}&$\alpha^{\ast}_{1}$ &$\alpha^{\ast}_{2}$ &$\alpha^{\ast}_{3}$ &$\alpha^{\ast}_{4}$  &$\alpha^{\ast}_{5}$ \\
 \cline{2-6}
{} & $0.2234(0.0404)$  &  $0.6613(0.2007)$ & $1.0869(0.2546)$& $0.9581(0.2774)$ &$1$\\\Xcline{2-6}{0.5pt}
\cline{2-6}
{}&$T^{\ast}_{r1}$ &$T^{\ast}_{r2}$ &$T^{\ast}_{r3}$ &$T^{\ast}_{r4}$&$T^{\ast}_{r5}$ \\
 \cline{2-6}
{}& $0.1049(0.0608)$  &  $0.5775(0.1096)$ & $0.8310(0.0667)$ &$0.7229(0.1432)$&$0.7491(0.1046)$ \\
\hline
\hline
\multirow{4}{*}{Bayes}&$\alpha^{\ast}_{1}$ &$\alpha^{\ast}_{2}$ &$\alpha^{\ast}_{3}$ &$\alpha^{\ast}_{4}$ &$\alpha^{\ast}_{5}$  \\
 \cline{2-6}
{} & $0.0219(0.0055)$  &  $0.4162(0.0428)$ & $0.9210(0.2150)$&$0.7418(0.3106)$ &$1$\\\Xcline{2-6}{0.5pt}
\cline{2-6}
{}&$T^{\ast}_{r1}$ &$T^{\ast}_{r2}$ &$T^{\ast}_{r3}$ &$T^{\ast}_{r4}$&$T^{\ast}_{r5}$  \\
 \cline{2-6}
{}& $0.0411(0.0125)$  &  $0.6620(0.1081)$ & $0.9351(0.0254)$ &$0.5595(0.1233)$&$0.7622(0.0907)$ \\
\hline

\end{tabular}
}
\end{center}
\footnotesize
(a) Evaluation of the methods, ``--'': not available, the best performance in each cell is bolded.

(b) Optimal weights and rejection thresholds are listed as mean(standard deviation).

\end{table*}



\subsection{Multi-Class Tasks}

The detailed results on the  multi-class data sets are shown from Table \ref{result_cardio} to Table \ref{result_pageblock}, including the performance evaluations and the values of the optimal parameters. Compared with the conventional classifications that have high error rates of the minority classes, \textit{SMOTE} is effective in reducing these errors. \textit{Cost-sensitive learning} classifies all instances to the class that has the minimum number of instances. Both \textit{Gmean\_no\_rej} and \textit{NI\_no\_rej} perform well with low error rates of the minority classes, high G-mean and high NI. \textit{Chow's reject} and \textit{Gmean\_rej} reject lots of instances from the minority classes; besides,  \textit{Gmean\_rej} rejects lots of instances from the majority class. 
On the whole, our \textit{NI\_rej} performs best with low error rates of the minority classes, a certain amount of reject rate, high G-mean, and the highest NI.

In summary, the  observations above suggest that:
\begin{enumerate}
\renewcommand{\labelenumi}{\theenumi.}
\item Within the CFL category, both \textit{SMOTE} and  G-mean based methods are  effective in the class imbalance problem. However, they are unable to process abstaining classifications.
\item Regarding to  \textit{Cost-sensitive learning}, it is feasible to apply the inverses of the class distribution ratios as the misclassification costs on binary class tasks. But on multi-class tasks, it may be ineffective. Moreover, the rejection costs are always hard to get. 
\item \textit{Chow's reject}   would perform poorly if the rejection thresholds are arbitrarily given.
\item NI based strategy is a good choice for both non-abstaining and abstaining classifications. It can produce reasonable solutions on the minority classes.
 \end{enumerate}

\section{Conclusion}
In this paper, we propose a new strategy of CFL to deal with the class imbalance problem. Based on the specific property of mutual information that can distinguish different error types and reject types, we seek to maximize it as a general rule for dealing with binary/multi-class classifications with/without abstaining. A unique feature is gained in abstaining classifications when information is unknown about errors and rejects. To our best knowledge, no other existing approach is applicable to this scenario. Moreover, we can derive the  ``{\it equivalent}''  costs for binary classifications. Generally, the ``{\it equivalent}'' costs will be changed accordingly to the distributions of the given data sets. Therefore, the present strategy provides an  ``{\it objective}''  reference for CSL if users want to adjust the costs. For better understanding ROC curves in binary classifications, graphical interpretations of the theoretical ROC curve plots are explained in terms of the related parameters, such as cost terms and rejection thresholds. Empirical study confirms the advantages of the proposed strategy in solving class imbalance problems. At the same time, we recognize the disadvantage of the work that it will add an extra computational cost over the existing approaches. This difficulty will form a future work for advancing the study.



%



\ifCLASSOPTIONcompsoc
  \section*{Acknowledgments}
\else
  \section*{Acknowledgment}
\fi

This work is supported in part by NSFC (No.
61075051) and the SPRP-CAS (No. XDA06030300). 

\ifCLASSOPTIONcaptionsoff
  \newpage
\fi

\end{document}